\documentclass{article}

\usepackage[nonatbib, preprint]{neurips_2020}

\usepackage[utf8]{inputenc} 
\usepackage[T1]{fontenc}    
\usepackage{url}            
\usepackage{booktabs}       
\usepackage{amsfonts}       
\usepackage{nicefrac}       
\usepackage{microtype}      

\usepackage{amssymb}
\usepackage{amsthm}

\usepackage{epsfig}
\usepackage{graphicx}
\usepackage{amsmath}
\usepackage{amssymb}
\usepackage{physics}
\usepackage{times}
\usepackage{helvet}
\usepackage{courier}

\usepackage{times}
\usepackage{xcolor}
\usepackage{bm}
\usepackage{multirow}
\usepackage{adjustbox}
\usepackage{xspace}
\usepackage{enumitem}
\usepackage{subcaption}
\usepackage{siunitx}
\usepackage{tabu}

\usepackage{mathtools}
\usepackage{comment}
\usepackage{titletoc}
\usepackage{hyperref} 

\newtheorem{Theorem}{Theorem}
\newtheorem{Proposition}{Proposition}
\newtheorem{Lemma}{Lemma}
\newtheorem{Corollary}{Corollary}[Theorem]
\newtheorem{definition}{Definition}
\newtheorem*{Claim}{Claim}

\newcommand{\rr}{\mathbb{R}}

\newcommand{\F}{\mathcal{F}}
\newcommand{\G}{\mathcal{G}_{\mathcal{F}}^{\mu}}
\newcommand{\y}{\mathcal{Y}}
\newcommand{\E}{\mathbb{E}}
\newcommand{\prob}{\mathbb{P}}
\newcommand{\sg}[1]{\textup{subG}(#1)}
\newcommand{\R}[1]{\mathbf{R}_{#1}}
\newcommand{\pr}{\textbf{p}_x}

\DeclareMathOperator*{\argmax}{arg\,max}

\allowdisplaybreaks[4]

\title{Higher-Order Certification for \\ Randomized Smoothing}

\author{Jeet Mohapatra$^1$ \And Ching-Yun Ko$^1$  \And Tsui-Wei Weng$^{1,2}$  \And Pin-Yu Chen$^{2}$  \And
Sijia Liu$^{2}$ \And Luca Daniel$^1$ \AND \\ ${}^1$ MIT \\
${}^2$ MIT-IBM Watson AI Lab, IBM Research 
}

\begin{document}
\maketitle
\begin{abstract}
     Randomized smoothing is a recently proposed defense against adversarial attacks that has achieved state-of-the-art provable robustness against $\ell_2$ perturbations. A number of publications have extended the guarantees to other metrics, such as $\ell_1$ or $\ell_\infty$, by using different smoothing measures. Although the current framework has been shown to yield near-optimal $\ell_p$ radii, the total safety region certified by the current framework can be arbitrarily small compared to the optimal.
 
    In this work, we propose a framework to improve the certified safety region for these smoothed classifiers without changing the underlying smoothing scheme. The theoretical contributions are as follows: 1) We generalize the certification for randomized smoothing by reformulating certified radius calculation as a nested optimization problem over a class of functions. 2) We provide a method to calculate the certified safety region using zeroth-order and first-order information for Gaussian-smoothed classifiers. We also provide a framework that generalizes the calculation for certification using higher-order information. 3) We design efficient, high-confidence estimators for the relevant statistics of the first-order information. Combining the theoretical contribution 2) and 3) allows us to certify safety region that are significantly larger than the ones provided by the current methods. On CIFAR10 and Imagenet datasets, the new regions certified by our approach achieve significant improvements on general $\ell_1$ certified radii and on the $\ell_2$ certified radii for color-space attacks ($\ell_2$ perturbation restricted to only one color/channel) while also achieving smaller improvements on the general $\ell_2$ certified radii.
     
    As discussed in the future works section, our framework can also provide  a way to circumvent the current impossibility results on achieving higher magnitudes of certified radii without requiring the use of data-dependent smoothing techniques.

 \end{abstract}
 
\section{Introduction}

Deep neural networks (DNNs) can be highly sensitive, i.e., small imperceptible input perturbations can lead to mis-classification \cite{szegedy2014intriguing,goodfellow2015explaining}. This poses a big problem for the deployment of DNNs in safety critical applications including aircraft control systems, video surveillance and self-driving cars, which require near-zero tolerance to lack of robustness. Thus, it is important to provide guarantees for the robustness of deep neural network models against multiple worst-case perturbations. Popular threat models are the $\ell_p$-norm-based attacks, where possible perturbations are constrained in an $\ell_p$-ball with respect to a given input $x$. To that end recent research efforts have focused on attack-agnostic robustness certification which, given an input $x$, provides a \textit{safety region} within which the model is guaranteed not to change its prediction. 

\textit{Randomized smoothing} is a recently-proposed defense \cite{Lecuyer2019Certified,li2018attacking,Cohen2019Certified} that has achieved state-of-the-art robustness guarantees. Given any classifier $f$, denoted as a \textit{base classifier}, randomized smoothing predicts the class that is ``most likely'' to be returned when noise is added to the input $x$. Thus, randomized smoothing acts as an operator that given a base classifier and a noise model, denoted as \textit{smoothing measure}, produces a new classifier, denoted as the \textit{smoothed classifier}. The smoothed classifiers thus produced are easily-certifiable with strong (even near-optimal for $\ell_2$) robustness guarantees under various $\ell_p$ norm threat models \cite{Lecuyer2019Certified,Li2019Certified,dvijotham2020framework}).

However, for any given threat model, the state-of-the-art robustness guarantees are achieved only for the smoothed classifiers obtained using very specific smoothing measures. Thus, the classifiers that attain state-of-the-art guarantees under one threat model might perform poorly under another (Gaussian-smoothed classifiers are optimal under $\ell_2$ but perform poorly under $\ell_1$; uniform noise-smoothed classifiers are state-of-the-art under $\ell_1$ but have poor performance under $\ell_2$). Moreover, some of the recent works~\cite{yang2020randomized, blum2020random, kumar2020curse} show that the existing framework, which uses only the zeroth order information (the function value of the smoothed classifier $g(x)$), is incapable of producing large certified radii for $\ell_p$ norms with high values of $p$.

Motivated by the two limitations above, we focus our attention on improving the certified safety region which is agnostic of threat models. To that end, we propose a general framework to provide a larger certified safety region by better utilizing the information derived from a hard-label classifier. In particular, we summarize our contributions as follows:
\begin{enumerate}
    \item We propose a general framework that calculates a certified safety region of a smoothed classifier $g$, around an input point $x$, by exploiting the estimated local properties (e.g. gradient, Hessian, etc.) of the classifier $g$ at $x$. 
    
    \item We give a threat-model-agnostic asymptotic-optimality result for smoothed classifiers obtained by using standard Gaussian as the smoothing measure, i.e. \textit{gaussian-smoothed} classifiers. Using Theorem \ref{thm: ident} in Section \ref{subsec: regular}, we show that theoretically it is possible to produce arbitrarily tight certificates for any classifier\footnote{For this result, the optimality assumes we abstain at points where the top-1 class has probability lower than 0.5.}. As a consequence, we see that the impossibility results for the existing framework cannot be extended to certificates obtained using higher-order information.
    
    \item We motivate and prove properties, like convexity (Proposition \ref{thm: convex}) and non-decreasing dependence on angle (Proposition \ref{thm: growth}), regarding the certified safety regions of gaussian-smoothed classifiers produced using the zeroth and first-order local information. Using these properties, we give formulas for calculating certified radii for gaussian-smoothed classifiers under $\ell_p$ threat models and their subspace variants with $p = 1, 2, \infty$. 
    
    \item We design new efficient estimators (see Table~\ref{tbl: estimators}) to provide high-confidence interval estimates of relevant first-order information about the classifier since the naive Monte-Carlo estimators have prohibitively high sample complexity.
\end{enumerate}

Finally, we use the 3rd and the 4th contributions above to empirically verify the effectiveness of the new certification framework in providing state-of-the-art certified accuracy for multiple threat models simultaneously. In particular, our proposed framework substantially boosts the certified accuracy for $\ell_1$ norm and subspace $\ell_2$ norm while maintaining (at times marginally improving) the state-of-the-art near-optimal results for $\ell_2$ norm. On the CIFAR10 dataset, our results for the $\ell_\infty$ norm also show improvement over the state-of-the-art bounds given by Gaussian smoothing.
\section{Background and Related Works}

\subsection{Related Works}
\textbf{Randomized Smoothing.} Randomized smoothing was initially introduced as a heuristic defense by \cite{liu2018towards} and \cite{xie2017mitigating}. Later, \cite{Lecuyer2019Certified} formulated it as a certification method using ideas from differential privacy, which was then improved by \cite{Li2019Certified} using Renyi divergence. For gaussian-smoothed classifiers, \cite{Cohen2019Certified} made the certified bounds worst-case-optimal in the context of certified $\ell_2$ norm radii by using the Neyman-Pearson Lemma, while authors of \cite{salman2019provably} combined the certification method with adversarial training to further improve the empirical results. 
Along another line of works, some extended existing certification methods to get better $\ell_p$ norm certified radii using different smoothing distributions, e.g. a discrete distribution for $\ell_0$ certificates \cite{lee2019tight}, the Laplace distribution for $\ell_1$ certificates \cite{teng2019adv}, and the generalized Gaussian distribution for $\ell_\infty$ certificates \cite{zhang2020soap}. Recently, \cite{yang2020randomized} proposed a general method for finding the optimal smoothing distribution given any threat model, as well as a framework for calculating the certified robustness for the smoothed classifier.

\textbf{Impossibility Results.} Recently, a number of works have shown that for $\ell_p$ norm threat models with large $p$, it is impossible to give a big  certified radius $\big(O(d^{\frac{1}{p}-\frac{1}{2}}\big)$ where $d$ is the input dimension) while retaining a high standard accuracy. In particular, the results on $\ell_\infty$ threat model given in ~\cite{blum2020random,kumar2020curse} and the results on $\ell_p$ (for sufficiently large $p$) threat models given in ~\cite{yang2020randomized} establish a certification/accuracy trade-off, which also exaggerates the need for an extended and generalized framework that breaks the confined trade-off and impossibility results.

\subsection{Preliminaries and Notations}
Generally, a classifier in machine learning is represented as a function $f$ that maps from the feature space $\rr^d$ to a probability vector over all the classes $\y$. We use $\F$ to denote the set of \textit{all} classifiers $f: \rr^d\mapsto \Delta^\y$, where $\Delta^\y$ is a probability simplex collecting all possible probability vectors with dimension $\y$. In this paper, we work under the restrictive black-box model, i.e., we assume we only have access to the class labels outputted by classifier $f$. For ease of use, we consider the output of $f$ to be the hard-thresholding of the probability vector, i.e. the output is a one-hot encoding vector. Given a black-box classifier $f \in \F$, randomized smoothing is a technique that creates a smoothed classifier $g$ by convolving $f$ (hereafter referred to as the \textit{base classifier}) with a \textit{smoothing measure} $\mu$, i.e., $g = f \star \mu$. We use $\G$ to denote the class of these smoothed functions and $\star$ denotes the convolution operator. Note that the smoothing measure $\mu$ is a probability measure and each smoothed function $g \in \G$ defines a mapping from $\rr^d$ to $\Delta^\y$. Thus, we have $\G \subset \F$.

\section{A General Framework for Randomized Smoothing}

Given a smoothing measure $\mu$, the certified safety region around a point $x$ is given as the subset of the space over which the classifier output is guaranteed not to change. In practice, this involves estimating some local information about $g$ around the point $x$ and giving the certified safety region as the common subset (intersection) of the safety region of all classifier $h \in \G$ whose local properties match the estimated local information. In the current literature, the only local information used is the function value $g(x)$~\cite{Cohen2019Certified, salman2019provably}; however, more local information of $g(x)$, such as the gradient or higher-order derivatives, can in fact be exploited for deriving better certified radii.
To illustrate the idea, we express the local information as the local constraints, and re-write the problem of finding the certified radius in the following: Let $H^x_i(g)$ be an estimate of some local property of $g$ at the input point $x$, and let there be $k$ such constraints, then the certified safety region $\mathbf{SR}\big(x\big)$ can be written as
\begin{equation}
    \begin{aligned}
        \label{eq: problemForm}
        \mathbf{SR}\big(x\big) &= \bigcap \Big\{ \mathbf{S}^h(x)   \Bigm|~ h \in \G \text{ and } H^x_i(h) = H^x_i(g), ~ 1 \leq i \leq k \Big\}, \\
        \mathbf{S}^h(x) &= \big\{ \delta \bigm| \argmax h(x + \delta) = \argmax h(x) \big\}
    \end{aligned}
\end{equation}
where $\mathbf{S}^h(x)$ gives the safety region for the function $h$. The existing state-of-the-art certificates~\cite{Cohen2019Certified} via randomized smoothing are a special case of this framework with $k = 1$ and $H^x_1(h) = (h(x))_A$ given $H^x_1(g) \geq p_A$, where $h(\cdot)_A$ denotes the $A^{th}$ component of the vector $h(\cdot)$ and $p_A$ is the lower bound of the estimated probability that $x$ belongs to the predicted class $A$.

\textbf{Reduction to Binary.} In existing literature, certificates for randomized smoothing based classifiers effectively reduce the multi-class classification to binary by reducing the problem to \textit{top-1 prediction vs other classes}. We follow the same idea to simplify our certification objectives by observing :
$$ h(x+\delta)_{\argmax(h(x))} > 0.5 \implies \argmax (h(x+\delta)) = \argmax (h(x)).$$
Assuming $\argmax(h(x)) = c$, we give by $\mathbf{S}^h_L(x) = \{ \delta \mid h(x + \delta)_c > 0.5 \}$ a set lower bound of $\mathbf{S}^h(x)$, i.e., $\mathbf{S}^h_L(x) \subset \mathbf{S}^h(x)$. Substituting $\mathbf{S}^h(x)$ in \eqref{eq: problemForm} by $\mathbf{S}^h_L(x)$ gives us a set lower bound $\mathbf{SR}_L(x)$ of $\mathbf{SR}\big(x\big)$ with the safety region subset:
\begin{equation*}
\begin{aligned}
    \mathbf{SR}_L(x) = \bigcap \Big\{ &\big\{ \delta \bigm | h(x + \delta)_c > 0.5 \big\} \Bigm | h \in \G ~ \& ~ H^x_i(h) = H^x_i(g), ~ 1 \leq i \leq k \Big\}
\end{aligned}
\end{equation*}

This set can be re-written as $\mathbf{SR}_L(x) = \{ \delta \mid \pr(x + \delta) > 0.5 \}$, where the probability function $\pr(z)$ given by the optimization problem:

\begin{align}
\label{eq: GenNP}
    \pr\big( z \big) = &\mathop{\boldsymbol\min}_{h \in \G}  h(z)_c \quad \textbf{ s.t.} \quad H^x_i(h) = H^x_i(g), ~ 1 \leq i \leq k.
\end{align}
\textbf{Our proposed framework.} Use Generalized Neymann Pearson Lemma \cite{chernoff1952generalization} to solve Equation \eqref{eq: GenNP} for $\pr(z)$. The safety region is then given as super-level set of $\pr$, i.e., $\mathbf{SR}_L(x) = \{ \delta \mid \pr(x + \delta) > 0.5 \}$.

\textbf{Discussion. } Under the proposed general framework for calculating certified radii, it is easy to see that adding more local constraints ($H^x_i$) in Equation \eqref{eq: GenNP} gives a bigger value of $\pr(z)$ for any $x, z$ which makes the super-level set of $\pr$, equivalently the certified safety region, bigger. In the following subsection, we study the properties of functions in $\G$ to get some motivation about which information might help us achieve a larger certified safety region. Then, we provide an example usage of this framework that exploits additional local higher-order information to give larger certified radii. The proofs of all theoretical results hereafter are supplemented in the appendix.
\subsection{Regularity Properties of $\mathbf{\G}$}
\label{subsec: regular}
For any black-box classifier $f$, the function map between the input pixel-space $\rr^d$ and the prediction $\Delta^\y$ is discontinuous. As a result, the higher-order derivatives might not always exist for the smoothed function $g$ for smoothing with general probability measures $\mu$. Therefore, it is crucial to establish conditions on the probability measure $\mu$ that guarantee the existence of higher-order derivatives for $g$ at all points in the pixel-space $\rr^d$. We give this in the following theorem: 
\begin{Theorem}
   If $\forall \alpha \in \mathbb{N}^d$, $\int_{\rr^d}\abs{D^{\alpha}_x\mu(y-x)}dy$ \footnote{$D^{\alpha}$ is the multi-variate differential operator, where $\alpha$ is a multi-index.} is finite, then $\G \subset \mathcal{C}^\infty$. Moreover, if $g \in \G$ is given as $g = f \star \mu$ for some $f \in \F$, then $$\nabla^i g(x) = \int_{\rr^d} f(y) (-1)^i\nabla^i \mu (y-x)  dy.$$
\end{Theorem}

\begin{Corollary}\label{cor: exist}
    When $\mu$ is the isotropic Gaussian distribution $\mathcal{N}(0, \sigma^2 \mathcal{I})$, then 
    $\G \subset \mathcal{C}^\infty$ and $\nabla^i g(x) = \int_{\rr^d} f(y) (-1)^i\nabla^i \mu (y-x)  dy$.
\end{Corollary}

Corollary \ref{cor: exist} guarantees the existence of higher-order derivatives for functions obtained by Gaussian smoothing and also provides a way of evaluating the higher-order terms using only the output of the base classifier $f$. Considering the fact that the truncated $m^{th}$ order Taylor expansion around $x$ gives better local approximations of the function with an increasing $m$ , we expect to get better approximations of the safety region by using the higher-order information about $g$ at $x$. Now in order to understand the limitations of this technique we consider the asymptotic limit where we have all the higher order terms of $g$ at $x$. We observe that:

\begin{Theorem}
\label{thm: ident}
    When $\mu$ is the isotropic Gaussian distribution $\mathcal{N}(0, \sigma^2 \mathcal{I})$, then $\forall g \in \G$, $g$ is a real analytic function with infinite radius of convergence, i.e., the Taylor series of $g$ around any point $w$ converges to the function $g$ everywhere.
\end{Theorem}

\textbf{Asymptotic-Optimality Remark:} As a consequence of Theorem \ref{thm: ident}, when $\mu$ is the isotropic Gaussian distribution, any function $h \in \G$ is uniquely identified by its Taylor expansion. Thus if we can accurately estimate all the higher-order information about $g$ at any point $x$, along with the fact that $g \in \G$, the feasible set in problem \eqref{eq: GenNP} will reduce to the singleton set containing only the function $g$ making $\pr(z) = g(z)_c$. This gives us the safety region $\mathbf{SR}_L(x) = \{ g(x)_c > 0.5\}$ which is the exact safety region under the binary, top-1 vs rest, relaxation (abstain when top-1 class has probability less than 0.5). Computing the exact safety region implies that we can get large certified $\ell_p$ radii for all $p$. Hence, the impossibility results shown in the recent works by Yang et.al \cite{yang2020randomized}, Blum et al. \cite{blum2020random} and Kumar et al. \cite{kumar2020curse} or any of their extensions can not hold for certification methods that use higher-order information. See \textsc{Appendix \ref{sec: Case_study}} for remarks on optimality of certificates obtained using only zeroth and first order information.
\subsection{Certification For Randomized Smoothing Using First-Order Information}
\label{subsec: firstRS}

For the rest of this paper, we focus on the certified region produced by solving the optimization problem described in Equation \eqref{eq: GenNP} using the zeroth and first-order local information of $g$ at a point $x$. The resulting first-order optimization problem is given as follows:
\begin{equation}
\label{eq: firstorder}
\begin{aligned}
\pr\big( z \big) = &\mathop{\boldsymbol\min}_{h \in \G}  h(z)_c \quad \textbf{ s.t.} \quad h(x) = y^{(0)}, \nabla h(x) = y^{(1)}
\end{aligned}
\end{equation}
where $y^{(0)}, y^{(1)}$ are the zeroth and first-order local information of $g$ at $x$ estimated using Corollary \ref{cor: exist}. In practice, it is only possible to give interval estimates for $y^{(0)}, y^{(1)}$. The following Theorem gives a lower bound for $\pr$ given interval estimates of $y^{(0)}, y^{(1)}$:

\begin{Theorem}[Lower Bound of $\pr(z)$]
\label{thm: direc}
    For a base classifier $f \in \F$, if $g = f \star \mu$, $\mu$ is the isotropic Gaussian distribution $\mathcal{N}(0, \sigma^2 \mathcal{I}), y^{(0)} = g(x)$, $y^{(1)} = \nabla g(x)$, then for any unit vector $v$ and any positive value of $r$, $\pr(x + \sigma rv)$ can be lower bounded by solving the following set of equations:
    \begin{align}
        \label{eq: probbnd}
        \int_{-\infty}^\infty \frac{1}{\sqrt{2\pi}}e^{-\frac{x^2}{2}} \Phi(c(x)) dx &= q  
    \end{align}
    \begin{subequations}
    \begin{minipage}{0.5\textwidth}
        \begin{align}
        \label{eq: ybnd}
            \int_{-\infty}^\infty
            \frac{1}{\sqrt{2\pi}}e^{-\frac{x^2}{2}} \frac{1}{\sqrt{2\pi}}e^{-\frac{c(x)^2}{2}} dx = m_2
        \end{align}
    \end{minipage}
    \begin{minipage}{0.5\textwidth}
        \begin{align}
        \label{eq: xbnd}
        \int_{-\infty}^\infty \frac{1}{\sqrt{2\pi}}xe^{-\frac{x^2}{2}} \Phi(c(x)) dx = m_1
        \end{align}
    \end{minipage}
    \end{subequations}
    with $q \leq y^{(0)}, m_1 \leq \sigma v^Ty^{(1)}, m_2 \leq \sigma \norm{y^{(1)} - v^Ty^{(1)}v}_2$, $c(x) := c_0 + c_1 x + c_2e^{rx}$, and $\Phi(z)$ being the CDF of the standard normal distribution. If the solution $(c_0, c_1, c_2)$ of above equations has $c_1 < 0$, then the lower bound of $\pr(x + \sigma rv)$ is instead given by solving Equations \eqref{eq: probbnd} to \eqref{eq: ybnd} with $c(x) := c_0 + c_2e^{rx}$.
\end{Theorem}

Given $\pr(z)$ the certified safety region can be given as the super-level set $\{ z \mid \pr(z) > 0.5 \}$. To supplement the above theorem, we point out that using only Equation \eqref{eq: probbnd} gives exactly the zeroth-order certification. Intuitively,  we see that unlike the zeroth-order certification that is calculated through equations independent of $v$, first-order certification uses additional Equations \eqref{eq: ybnd}, \eqref{eq: xbnd} that depend on $v$, breaking the symmetry and admitting non-isotropic certified radius bounds. 
Referring to Theorem~\ref{thm: direc}, we observe that $\pr(x + rv)$ only depends on $g(x), \norm{\nabla g(x)}_2$ and angle between $v$ and $\nabla g(x)$. Thus, the first order certified region has a cylindrical symmetry with the axis along the vector $y^{(1)}$. 
\begin{figure}[t]
    \centering
    \includegraphics[width=1.0\linewidth]{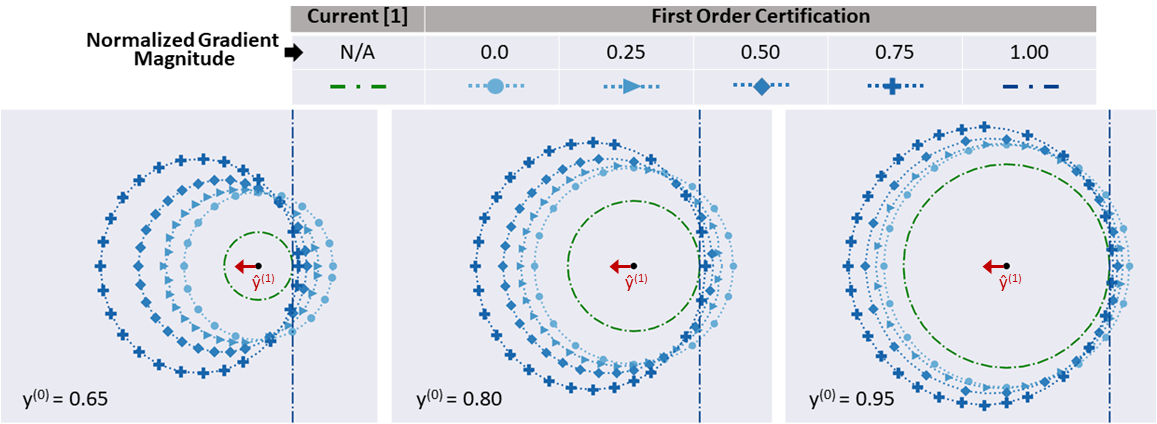}
     \caption{A longitudinal slice of the certified safety region around a point under various values of $y^{(0)}, y^{(1)}$ keeping direction of $y^{(1)}$ fixed along the negative x axis. Normalized gradient magnitude is given as $\norm{y^{(1)}}_2 /\max_{g(x) = y^{(0)}} \norm{\nabla g(x)}_2$.The green region is certified using only zeroth order information and the blue regions using both zeroth and first order information.}
     \label{fig: safetyRegion}
\end{figure}

In Figure \ref{fig: safetyRegion}, we compare a longitudinal slice of the safety region around a point $x$ calculated using both zeroth and first-order information to the longitudinal slice of the safety region obtained using only the zeroth-order information. We use these figures to deduce some properties of these safety regions:

\textbf{i)} The relative improvement in the size of the safety region is the largest for low values of $y^{(0)}$ and get successively smaller with an increasing $y^{(0)}$.

\textbf{ii)}  Given a fixed value of $y^{(0)}$, the volume of the certified safety region grows to infinity (the half-space given by the dash line) as the magnitude of $\norm{y^{(1)}}_2$ increases to its maximum possible value.

\textbf{iii)} The certified safety region is convex and for any given values of $y^{(0)}$ and $\norm{y^{(1)}}_2$, the directional certified radius is highest along the direction of the $y^{(1)}$ and gets successively lower as we rotate away from it (the angle between $v$ and $y^{(1)}$ increases), to the lowest in the direction opposite to the $y^{(1)}$.

We formalize the third observation into the following propositions, which allows us to calculate the certified radii for the various threat models discussed in the rest of this section.

\begin{Proposition}
\label{thm: convex}
    The certified safety region, $\mathbf{SR}_L(x)$, calculated using the zeroth and first-order local information is convex, i.e., if $x_1, x_2 \in \mathbf{SR}_L(x)$ then $\frac{x_1 + x_2}{2} \in \mathbf{SR}_L(x)$.
\end{Proposition}

Using Prop \ref{thm: convex}, we see that along any vector $v$ if for some $R$, $x + Rv \in \mathbf{SR}_L(x)$, then forall $0 \leq r \leq R$, $x + rv \in \mathbf{SR}_L(x)$. Thus, along any direction $v$, we can define a directional robustness $\R{v}$ such that forall $0 \leq r < \R{v}$, $x + rv \in \mathbf{SR}_L(x)$ and forall $r \geq R, x + rv \not \in \mathbf{SR}_L(x)$. It is easy to see that $\R{v}$ can be given as the solution to the equation $r \geq 0, \pr(x + rv) = 0.5$.
\begin{Proposition}
\label{thm: growth}
    For any given value of $y^{(0)}, y^{(1)}$, the directional robustness along $v$, $\R{v}$, given by the first-order certification method is a non-increasing function of the angle between $v$ and $y^{(1)}$\mbox{, i.e., $\cos^{-1}\Big(\frac{v^Ty^{(1)}}{\norm{v}_2\norm{y^{(1)}}_2}\Big)$}.
\end{Proposition}
 
\textbf{(I) Certified $\ell_2$ Norm Radius: }By Proposition \ref{thm: growth}, we have that for first-order certification, the directional robustness along all directions $v$ is at least as big as the directional robustness along the direction $-y^{(1)}$, implying that the $\ell_2$ norm certified radius is equal to $\mathbf{R}_{-y^{(1)}}$. Thus,we have
\begin{Corollary}[Certified $\ell_2$ Norm Radius]
\label{cor: L2rad}
    For a base classifier $f \in \F$, if $g = f \star \mu$, where $\mu$ is the isotropic Gaussian distribution $\mathcal{N}(0, \sigma^2 \mathcal{I}), y^{(0)} = g(x)$, $y^{(1)} = \nabla g(x)$, the $\ell_2$ norm radius $R$ is given as $R = \sigma r$, where $(r, w_1, w_2)$ is the solution of the system of equations:
    \begin{equation}
    \label{eq: L2sol}
        \Phi(w_1 - r) - \Phi(w_2 - r) = 0.5
    \end{equation}
    \begin{subequations}
    \begin{minipage}{0.5\textwidth}
        \begin{equation}
        \Phi(w_1) - \Phi(w_2) = q \label{const: prob}
        \end{equation}
    \end{minipage}
    \begin{minipage}{0.5\textwidth}
        \begin{equation}
        \frac{1}{\sqrt{2\pi}}e^{-\frac{w_2^2}{2}} -  \frac{1}{\sqrt{2\pi}}e^{-\frac{w_1^2}{2}} = m_1 \label{const: gradnorm}
        \end{equation}
    \end{minipage}
    \end{subequations}
    
    with $q \leq y^{(0)}$ and $m_1 \geq \sigma\norm{y^{(1)}}_2$. 
\end{Corollary} 

Corollary \ref{cor: L2rad} gives the same result as Cohen et al. \cite{Cohen2019Certified} if we remove constraint \eqref{const: gradnorm} and consider the minimum value of $r$ produced by Equation \eqref{eq: L2sol} over all possible pairs of $w_1, w_2$ satisfying equation \eqref{const: prob} . As a result the certified $\ell_2$ norm radius given by Corollary \ref{cor: L2rad} is  always greater than or equal to the radius given by the existing framework with equality holding only when the classifier is linear.

\textbf{(II) Certified $\ell_1, \ell_\infty$ Norm Radius: } Using Proposition \ref{thm: convex} we see that the minimum $\ell_1$ norm radius is along one of the basis vectors and using Proposition \ref{thm: growth} we see that the minimum directional robustness must be along the basis vector with the biggest angle with $y^{(1)}$. Thus, the projection of $y^{(1)}$ along this can be given as $-\norm{y^{(1)}}_\infty$. 

\begin{Corollary}[Certified $\ell_1$ Norm Radius]
\label{cor: L1rad}
    For a base classifier $f \in \F$, if $g = f \star \mu$, where $\mu$ is the isotropic Gaussian distribution $\mathcal{N}(0, \sigma^2 \mathcal{I}), y^{(0)} = g(x)$, $y^{(1)} = \nabla g(x)$, the $\ell_1$ norm radius $R$ is obtained by solving $\pr(x + Rv) = 0.5$, where $\pr(x + Rv)$ is given by solving the problem in Theorem \ref{thm: direc} with \mbox{$m_1 \leq -\sigma\norm{y^{(1)}}_\infty$}, $m_2 \leq \sigma \sqrt{\norm{y^{(1)}}^2_2 - \norm{y^{(1)}}_\infty^2}$.
\end{Corollary}

Similarly, the minimum $\ell_\infty$ norm radius must ocuur along a vector of all 1's and -1's and using Proposition \ref{thm: growth} we see it must be along the vector of all 1's and -1's with the biggest angle with $y^{(1)}$. Thus, the projection of $y^{(1)}$ along this can be given as $-\norm{y^{(1)}}_1$.

\begin{Corollary}[Certified $\ell_\infty$ Norm Radius]
\label{cor: LIrad}
    For a base classifier $f \in \F$, if $g = f \star \mu$, where $\mu$ is the isotropic Gaussian distribution $\mathcal{N}(0, \sigma^2 \mathcal{I}), y^{(0)} = g(x)$, $y^{(1)} = \nabla g(x)$, the $\ell_\infty$ norm radius $R$ is obtained by solving $\pr(x + Rv) = 0.5$, where $\pr(x + Rv)$ is given by solving the problem in Theorem \ref{thm: direc} with \mbox{$m_1 \leq -\frac{\sigma}{\sqrt{d}}\norm{y^{(1)}}_1, m_2 \leq \frac{\sigma}{\sqrt{d}} \sqrt{d\norm{y^{(1)}}^2_2 - \norm{y^{(1)}}_1^2}$}.
\end{Corollary}

\textbf{(IV) Certified $\ell_p$ Norm Radius over a Subspace: }
\begin{definition}[Subspace Certified $\ell_p$ Norm Radius]
 Given subspace $S$, we define the $\ell_p$ certified radius in the subspace as follows:
 \begin{align*}
     \max_R R
        \quad \textbf{ s.t. } \forall ~ \delta \in S, \norm{\delta}_p \leq R; ~  \argmax(g(x + \delta)) = \argmax(g(x)).
 \end{align*}
\end{definition}

 Going beyond $\ell_p$ norm threat models, we give the subspace $\ell_p$ norm threat model that allows us to, among other things, measure the sensitivity of a deep learning model over a subset of the pixels instead of the entire image. Using Propositions \ref{thm: convex} and \ref{thm: growth}, we are able to extend our results to give :

\begin{Corollary}[Subspace Certified $\ell_p$ norm radius]
\label{cor: subspace}
    For a base classifier $f \in \F$, if $g = f \star \mu$, where $\mu$ is the isotropic Gaussian distribution $\mathcal{N}(0, \sigma^2 \mathcal{I}), y^{(0)} = g(x)$, $y^{(1)} = \nabla g(x)$, and a subspace $S$ with orthogonal projection matrix $P_S$, for $p= 1, 2, \infty$ the subspace $\ell_p$ norm certified radius $R$ is obtained by solving $\pr(x + Rv) = 0.5$, where $\pr(x + Rv)$ is given by solving the problem in Theorem \ref{thm: direc} with $m_1 \leq -\sigma\norm{P_Sy^{(1)}}_{p'}, m_2 \leq \sigma \sqrt{\norm{y^{(1)}}_2^2 - \norm{P_Sy^{(1)}}_{p'}^2}$ and $\|\cdot\|_{p'}$ is the dual norm of $\|\cdot\|_{p}$.
\end{Corollary}

\section{Numerical Estimation of First-Order Information}
\begin{table*}[t]
    \centering
    \caption{Estimators to calculate the different norm value of $\nabla g(x)$. (\textbf{*} newly designed estimators)}
    {
    \begin{tabular}{c|ccc}
    \toprule
    & $\ell_1$ & $\ell_2 $ & $\ell_\infty$ \\
    \toprule
     & &$\epsilon_u = \sqrt{\frac{-k(n_1 + n_2)\log{\frac{\alpha}{2}}}{2n_1n_2(X_{n_1}^TY_{n_2} + t)}}$ &  \\ 
     
    constants & $t = \sqrt{\frac{2kd(d\log{2} - \log{\alpha})}{n_1 + n_2}}$ & $t = \sqrt{-k^2\frac{\sqrt{2}d}{n_1n_2}\log{\frac{\alpha}{2}}} $ & $t = \sqrt{\frac{2k(\log{2d} - \log{\alpha})}{n_1 + n_2}}$ \\
    
    &  & $\epsilon_l = \sqrt{\frac{-k(n_1 + n_2)\log{\frac{\alpha}{2}}}{2n_1n_2(X_{n_1}^TY_{n_2} - t)}}$ &  \\
    
    \midrule
    
    upper bound & $\norm{\frac{n_1X_{n_1} + n_2 Y_{n_2}}{n_1 + n_2}}_1 + t$ & $\frac{\sqrt{X_{n_1}^TY_{n_2} + t}}{\sqrt{1 + \epsilon_u^2} - \epsilon_u}$ \textbf{(*)} & $\norm{\frac{n_1X_{n_1} + n_2 Y_{n_2}}{n_1 + n_2}}_\infty + t$ \\
    
    lower bound & $\norm{\frac{n_1X_{n_1} + n_2 Y_{n_2}}{n_1 + n_2}}_1 - t$ & $\frac{\sqrt{X_{n_1}^TY_{n_2} - t}}{\sqrt{1 + \epsilon_l^2} + \epsilon_l}$ \textbf{(*)} & $\norm{\frac{n_1X_{n_1} + n_2 Y_{n_2}}{n_1 + n_2}}_\infty - t$ \\
    \bottomrule
    \end{tabular}
    }
    \label{tbl: estimators}
\end{table*}
In the previous section a concrete set of methods has been proposed to calculate the first-order certificate for randomized smoothing assuming we have access to the first-order information $y^{(1)}$. Therefore in this section we complete the framework by giving ways to get high confidence estimates of vector $y^{(1)}$. Recalling that the hard-label classifier $f$ is not continuous and hence not differentiable, we need to use the zeroth order information about $f$ to approximate $y^{(1)}$ , i.e., use the result from Corollary \ref{cor: exist} :  $y^{(1)} = \E_{y \sim \mu} [\frac{y}{\sigma^2} f(x+y)]$. As the estimator $\frac{y}{\sigma^2}f(x+y)$ is noisy \cite{salman2019provably}, approximating the expectation using Monte-Carlo sampling is hard, i.e., the number of samples required to achieve a non-trivial high-confidence estimate of $y^{(1)}$ scales linearly with the input dimension. In practice, this translates to a sample complexity of 10 billion for CIFAR and near a trillion for Imagenet. 

Although estimating the vector $y^{(1)}$ is hard, we observe from Corollary \ref{cor: L1rad}, \ref{cor: L2rad}, \ref{cor: LIrad}, \ref{cor: subspace} that we don't need the whole vector $y^{(1)}$ but only $\norm{y^{(1)}}_2$, $\norm{y^{(1)}}_\infty$ for $\ell_1$ certified radius; $\norm{y^{(1)}}_2$ for $\ell_2$ certified radius; $\norm{y^{(1)}}_1$, $\norm{y^{(1)}}_2$ for $\ell_\infty$ certified radius; and $\norm{P_Sy^{(1)}}_p$, $\norm{y^{(1)}}_2$ for the subspace $\ell_p$ certified radius ($p =1, 2, \infty$). Most of these statistics can be estimated much more efficiently due to the following observation :
\begin{Theorem}
\label{thm: subgauss}
    Given a black-box classifier $f$ and the random vector $z = w(f(x+w)_c - \frac{1}{2}) $ where $w \sim \mathcal{N}(0, \sigma^2 \mathbf{I})$, we have that $z - \sigma^2y^{(1)}$ is a sub-gaussian random vector with parameter $k = \sigma^2(\frac{1}{4} + \frac{3}{\sqrt{8\pi e}})$. For convenience, we do some abuse of notation to denote this as $(z - \sigma^2y^{(1)}) \sim \sg{k}$.
\end{Theorem} 
Leveraging the properties of sub-gaussian random variables, we establish the following general result:

\begin{Theorem}
\label{thm: estimator}
    For any $\alpha \geq 2e^{-\frac{d}{16}}$, if we have two random vectors $X, Y$ such that $(X - \beta) \sim \sg{k_1}$ and $(Y - \beta) \sim \sg{k_2}$, then we can show that using
    $t = \sqrt{-\sqrt{2}k_1k_2d\log{\frac{\alpha}{2}}} $, $\epsilon_u = \sqrt{\frac{-(k_1 + k_2)\log{\frac{\alpha}{2}}}{2(X^TY + t)}}$, and $\epsilon_l = \sqrt{\frac{-(k_1 + k_2)\log{\frac{\alpha}{2}}}{2(X^TY - t)}}$ renders
    \begin{equation*}
        \begin{aligned}
            \prob \bigg(\norm{\beta}_2 \leq \frac{\sqrt{X^TY + t}}{\sqrt{1 + \epsilon_u^2} - \epsilon_u}\bigg) \geq 1 - \alpha, \quad  \prob \bigg(\norm{\beta}_2 \geq \frac{\sqrt{X^TY - t}}{\sqrt{1 + \epsilon_l^2} + \epsilon_l}\bigg) \geq 1 - \alpha.
        \end{aligned}
    \end{equation*}
\end{Theorem}

Now, let $X = X_{n_1}, Y = Y_{n_2}$ be the empirical average of $n_1, n_2$ independent samples of the random variable $z$. Then, using Theorem \ref{thm: subgauss} and \ref{thm: estimator} allows us to give the estimators illustrated in Table \ref{tbl: estimators}. However, we see that estimation of $\norm{y^{(1)}}_1$ still scales linearly with $d$ making it impractical. Thus, the $\ell_\infty$ norm certified radius is still given as $\frac{1}{\sqrt{d}}$ of the $\ell_2$ norm certified radius. 

Finally, for estimating the $\ell_p$ norm ($p = 1, 2, \infty$) over a subspace $S$ we show that we can use the corresponding estimators from Table \ref{tbl: estimators} except with $X_{n_1}^S = P_S X_{n_1}$, $Y_{n_2}^S = P_S Y_{n_2}$ and $d = d_S$ where $P_S$ is the orthogonal projection matrix for $S$ and $d_S$ is the dimension of $S$.

\begin{figure}[t]
    \centering
    \begin{subfigure}[t]{0.48\linewidth}
    \includegraphics[width=\textwidth]{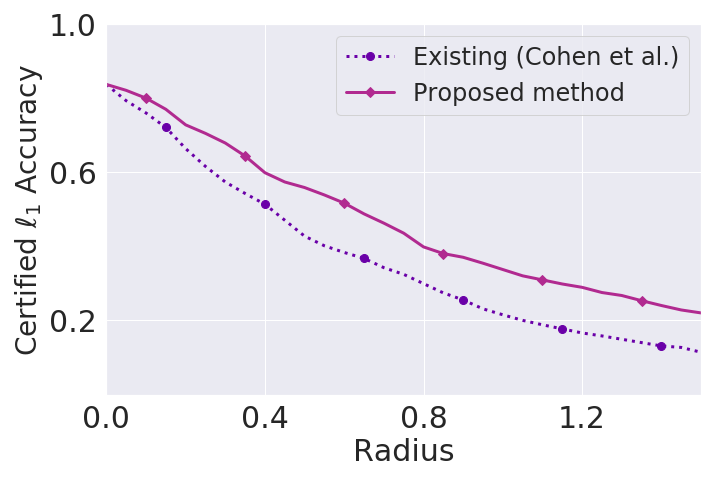}
    \caption{Certified $\ell_1$ radius}
    \label{fig: L1CertAcc}
    \end{subfigure}
    \begin{subfigure}[t]{0.48\linewidth}
    \includegraphics[width=\textwidth]{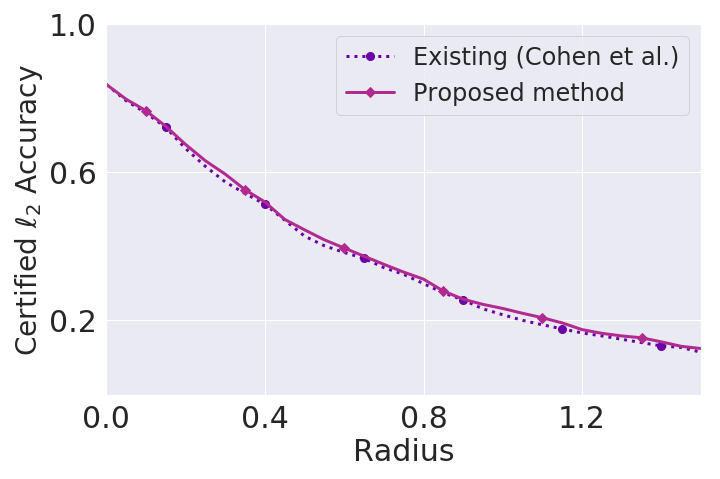}
    \caption{Certified $\ell_2$ radius}
    \label{fig: L2CertAcc}
    \end{subfigure}
    \begin{subfigure}[t]{0.48\linewidth}
    \includegraphics[width=\textwidth]{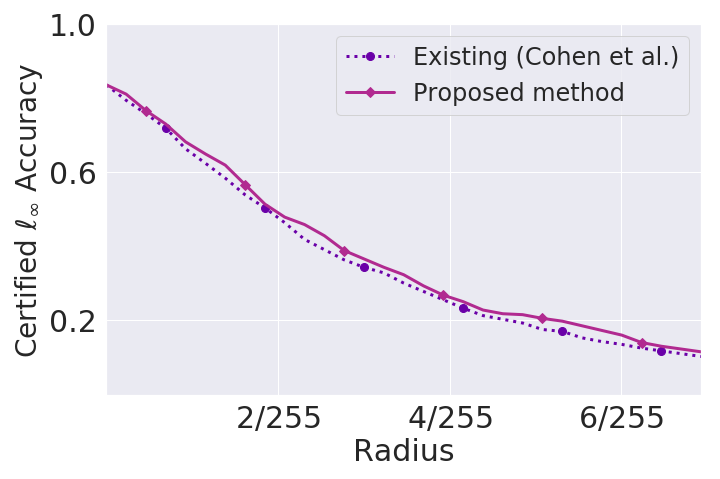}
    \caption{Certified $\ell_\infty$ radius}
    \label{fig: LICertAcc}
    \end{subfigure}
    \begin{subfigure}[t]{0.48\linewidth}
    \includegraphics[width=\textwidth]{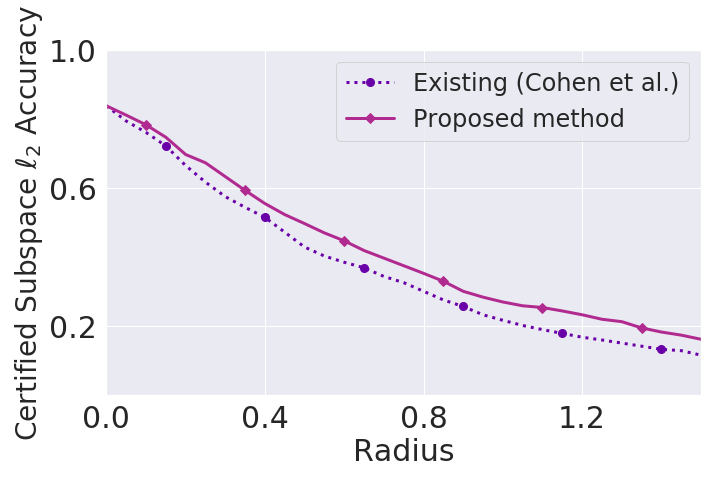}
    \caption{Certified Subspace $\ell_2$ radius}
    \label{fig: L2SubCertAcc}
    \end{subfigure}
    \caption{Increase in Certified Accuracy for CIFAR10 seen under various threat models}
\end{figure}

\section{Experimental Results}

We empirically study the performance of the new certification schemes on standard image classification datasets, CIFAR10 and Imagenet. We reuse the models given by Cohen et al. \cite{Cohen2019Certified} and calculate the certified accuracy at radius $R$ by counting the samples of the test set that are correctly classified by the smoothed classifier $g$ with certified radii of at least $R$. For both our proposed certificate and the baseline certificate~\cite{Cohen2019Certified}, we use a failure probability of $\alpha = 0.001$ and $N = 200,000$ samples for CIFAR10 and $N=1,250,000$ samples for Imagenet. For $\ell_\infty$ radius we require a lot more samples to get better results as our current estimator is too noisy. In order to show the possibility presented by our approach, we also provide results for the $\ell_\infty$ certified radius for CIFAR10 estimated using $4,000,000$ samples. In our plots, we present, for each threat model the upper envelopes of certified accuracies attained over the range of considered $\sigma \in \{0.12, 0.25, 0.50, 1.00\}$. Further experiments (including all Imagenet experiments) are given in the appendix.
\subsection{Certified Accuracy}

As expected we see from Figure \ref{fig: L2CertAcc} that the new framework gives only marginal improvement over the $\ell_2$ radius certified by existing methods. This follows from the fact that the existing methods already produce near-optimal certified $\ell_2$ radii. However, certifying a significantly bigger certified safety region allows us to give significant improvements over the certified $\ell_1$ radius, the $\ell_\infty$ radius and the subspace $\ell_2$ radii (the subspace considered here is the red channel of the image, i.e., we only allow perturbations over red component of the RGB pixels of the image). We note here that considering the performance for individual threat models, the $\ell_1$ norm certified accuracy given here is still smaller than the existing state-of-the-art. However, the current state-of-the-art for $\ell_1$ certified radius uses the uniform smoothing distribution which performs quite poorly for $\ell_2$ norm radius. If we consider multiple threat models simultaneously the proposed method gives the best joint performance. Similar findings are also reported for Imagenet experiments (given in the appendix). 

Moreover, we see that for the certified $\ell_\infty$ radius the new method gives an improvement over the certified $\ell_\infty$ radius provided by existing methods for Gaussian noise. As Gaussian noise was shown to achieve the best possible certified $\ell_\infty$ radius for existing methods, this provides an improvement over the best possible $\ell_\infty$ radius certifiable using existing methods. Although the improvements are small in magnitude, we believe they can be improved upon by using tighter estimators for $\norm{y^{(1)}}_1$.

\section{Conclusion and Future Work}
In this work, we give a new direction for improving the robustness certificates for randomized-smoothing-based classifiers. We have shown that even in the black-box (hard-label classifier) setting, leveraging more information about the distribution of the labels among the sampled points allows us to certify larger regions and thus guarantee large certified radii against multiple threat models simultaneously. We have shown this to hold theoretically and also demonstrated it on CIFAR and Imagenet classifiers. Additionally, for gaussian smoothing, the proposed framework gives a way to circumvent the recently-proposed impossibility results and also promises a threat-model-agnostic asymptotic-optimality result. However, we note that the first-order smoothing technique given in this paper is only a proof-of-concept to show it is possible to better leverage local information to certify larger safety regions without changing the smoothing measure. In future, this work could be extended to derive and use, for any given threat model, the best local information to exploit in order to improve the certificates for that threat model.
\clearpage

\section*{Broader Impact}

In recent years, machine learning and intelligent systems have started to be widely adopted into everyday life, including several safety-critical applications. The wide-spread use of these systems requires them to be held to a higher level of scrutiny. One such requirement is ``robustness'', i.e., the systems' prediction should not be too sensitive to small input perturbations. In general, undefended/vanilla deep learning models have been found to be extremely sensitive to small imperceptible perturbations. This in turn has given rise to many defense techniques, able to overcome ``existing'' threats to robustness. However most of these models have later proved to be ineffective against newer threats. In turn, this has then given rise to a class of techniques which provide some mathematical guarantees about the robustness of their predictions. In this paper, we  extended one such framework that provides some mathematically provable guarantees against any  ``single individual'' threat model chosen from a large class of threat models. Specifically, we provided a way to construct models with some mathematically provable  guarantees against ``multiple simultaneous'' threat models. 

The {\it benefits} of this contribution include providing a more holistic picture of the robustness guarantees for deployed models. This can hopefully bring us a step closer to trustworthy AI. Since this work helps building models that give some guarantees on their behavior, we hope it would also lead to the wider adoption of deep learning models. Moreover, we hope that the guarantees given by these models would allow people to have some improved sense of security when using deep-learning-based products. Considering intensive applications such as product matching, categorization in assembly lines or video surveillance that require long periods of hard effort, robust models will give more reliable and efficient means of achieving the task and considerably reducing the burden on humans.

As for the potentially {\it negative} effects of this contribution, we feel that any progress towards robustness certification could easily become a double-edged sword. This is because, if adopted blindly (i.e. specifically without extremely careful attention to the types of guarantees provided and most importantly those NOT provided), it may give also the {\bf false} sense of security that current deep-learning systems are already ready for deployment. These robustness concerns are one of the major bottlenecks for the adoption of deep-learning models in safety-critical applications such as self-driving car and air traffic control systems. hence the ability to provide ``some'' robustness guarantees might result in wide premature adoption  of deep learning models in such applications. We would like to candidly admit to developer and user readers of this paper that the presence of some robustness guarantees for a model does not mean we understand the model or that it is completely safe to deploy it. To the contrary the authors of this paper believe that we are still quite far from such safe adoption scenario. Further issues lie also beyond robustness. A negative effect of this publication and publication like this on robustness, is might give the highly incorrect impression that with robustness one can feel safe in deploying AI systems in society. On the contrary we would also like to candidly admit and remind the readers that deep-learning-based models suffer from problems such as lack of accuracy guarantees for out-of-distribution data , lack of  fairness, lack of explain-ability and many others that MUST to be solved before AI systems are viable for real-world applications. More specifically about the lack of accuracy guarantees for out-of-distribution data: the robustness of a model does not necessarily mean its prediction is always accurate. Robustness and accuracy are indeed two disjoint concepts. While it is well-known that an accurate prediction might not be robust, it is also essential to keep in mind that a robust prediction need not always be accurate. In other words, some models may generate a prediction that is indeed highly robust (i.e. does not change upon perturbation) but consistently incorrect! As a result, in applications such as air traffic control systems, the models might display extremely bad behaviour because of situations not present in its training, while the presence of some robustness guarantees might give a false sense of security to an inexperience user (e.g. not familiar with the admittedly subtle  mathematical intricacies of different threat models), that these systems are completely fault-tolerant. 

In conclusion, we would like to remark that although this paper helps to take a step towards building more trustworthy AI, that goal is indeed still quite far-off. 
\clearpage
\bibliographystyle{ieeetr}
\bibliography{style/neurips20}

\begin{thebibliography}{10}

\bibitem{szegedy2014intriguing}
C.~Szegedy, W.~Zaremba, I.~Sutskever, J.~Bruna, D.~Erhan, I.~Goodfellow, and
  R.~Fergus, ``Intriguing properties of neural networks,'' {\em ICLR}, 2014.

\bibitem{goodfellow2015explaining}
I.~Goodfellow, J.~Shlens, and C.~Szegedy, ``Explaining and harnessing
  adversarial examples,'' in {\em ICLR}, 2015.

\bibitem{Lecuyer2019Certified}
M.~{Lecuyer}, V.~{Atlidakis}, R.~{Geambasu}, D.~{Hsu}, and S.~{Jana},
  ``Certified robustness to adversarial examples with differential privacy,''
  in {\em 2019 IEEE Symposium on Security and Privacy (SP)}, pp.~656--672,
  2019.

\bibitem{li2018attacking}
Y.~Li, X.~Bian, and S.~Lyu, ``Attacking object detectors via imperceptible
  patches on background,'' {\em arXiv preprint arXiv:1809.05966}, 2018.

\bibitem{Cohen2019Certified}
J.~Cohen, E.~Rosenfeld, and Z.~Kolter, ``Certified adversarial robustness via
  randomized smoothing,'' in {\em Proceedings of the 36th International
  Conference on Machine Learning} (K.~Chaudhuri and R.~Salakhutdinov, eds.),
  vol.~97 of {\em Proceedings of Machine Learning Research}, (Long Beach,
  California, USA), pp.~1310--1320, PMLR, 09--15 Jun 2019.

\bibitem{Li2019Certified}
B.~Li, C.~Chen, W.~Wang, and L.~Carin, ``Certified adversarial robustness with
  additive noise,'' in {\em NeurIPS}, 2019.

\bibitem{dvijotham2020framework}
K.~D. Dvijotham, J.~Hayes, B.~Balle, Z.~Kolter, C.~Qin, A.~Gy{\"o}rgy, K.~Xiao,
  S.~Gowal, and P.~Kohli, ``A framework for robustness certification of
  smoothed classifiers using f-divergences.,'' in {\em ICLR}, 2020.

\bibitem{yang2020randomized}
G.~Yang, T.~Duan, E.~Hu, H.~Salman, I.~Razenshteyn, and J.~Li, ``Randomized
  smoothing of all shapes and sizes,'' {\em arXiv preprint arXiv:2002.08118},
  2020.

\bibitem{blum2020random}
A.~Blum, T.~Dick, N.~Manoj, and H.~Zhang, ``Random smoothing might be unable to
  certify $l_\infty$ robustness for high-dimensional images,'' {\em arXiv
  preprint arXiv:2002.03517}, 2020.

\bibitem{kumar2020curse}
A.~Kumar, A.~Levine, T.~Goldstein, and S.~Feizi, ``Curse of dimensionality on
  randomized smoothing for certifiable robustness,'' {\em arXiv preprint
  arXiv:2002.03239}, 2020.

\bibitem{liu2018towards}
X.~Liu, M.~Cheng, H.~Zhang, and C.-J. Hsieh, ``Towards robust neural networks
  via random self-ensemble,'' in {\em Proceedings of the European Conference on
  Computer Vision (ECCV)}, pp.~369--385, 2018.

\bibitem{xie2017mitigating}
C.~Xie, J.~Wang, Z.~Zhang, Z.~Ren, and A.~Yuille, ``Mitigating adversarial
  effects through randomization,'' {\em arXiv preprint arXiv:1711.01991}, 2017.

\bibitem{salman2019provably}
H.~Salman, J.~Li, I.~Razenshteyn, P.~Zhang, H.~Zhang, S.~Bubeck, and G.~Yang,
  ``Provably robust deep learning via adversarially trained smoothed
  classifiers,'' in {\em Advances in Neural Information Processing Systems},
  pp.~11289--11300, 2019.

\bibitem{lee2019tight}
G.-H. Lee, Y.~Yuan, S.~Chang, and T.~S. Jaakkola, ``Tight certificates of
  adversarial robustness for randomly smoothed classifiers,'' in {\em Advances
  in Neural Information Processing Systems}, 2019.

\bibitem{teng2019adv}
J.~Teng, G.-H. Lee, and Y.~Y. J., ``$\ell 1$ adversarial robustness
  certificates: a randomized smoothing approach,'' 2019.

\bibitem{zhang2020soap}
D.~Zhang*, M.~Ye*, C.~Gong*, Z.~Zhu, and Q.~Liu, ``Filling the soap bubbles:
  Efficient black-box adversarial certification with non-gaussian smoothing,''
  2020.

\bibitem{chernoff1952generalization}
H.~Chernoff and H.~Scheffe, ``A generalization of the neyman-pearson
  fundamental lemma,'' {\em The Annals of Mathematical Statistics},
  pp.~213--225, 1952.

\bibitem{boyd2004convex}
S.~Boyd, S.~P. Boyd, and L.~Vandenberghe, {\em Convex optimization}.
\newblock Cambridge university press, 2004.

\bibitem{wendel1948note}
J.~Wendel, ``Note on the gamma function,'' {\em The American Mathematical
  Monthly}, vol.~55, no.~9, pp.~563--564, 1948.

\end{thebibliography}
\newpage
\appendix

\section*{\centering \Large{Appendix}}
\setcounter{Theorem}{0}
\setcounter{Lemma}{0}
\setcounter{Proposition}{0}

\section{Mathematical Preliminaries}

In this section, we provide a brief introduction to the mathematical tools used in proving the theorems given in this paper :

\paragraph{Log-Concave Functions \cite{boyd2004convex}} A non-negative function $f : \rr^n \rightarrow \rr^{+}$ is log-concave if it satisfies the inequality
\[ f(\theta x+(1-\theta )y)\geq f(x)^{\theta }f(y)^{1-\theta}\]
for all $x,y \in \rr$ and $0 < \theta < 1$. Some popular examples of log-concave functions are the Gaussian pdf and $0-1$ indicator functions of convex sets. Log-concave functions satisfy the following properties:
\begin{itemize}
    \item Log-concave functions are also quasi-concave.
    \item If $f,g$ are both log-concave functions then the convolution $f\star g$ is also log-concave.
\end{itemize}

\paragraph{Quasi-Concave Functions \cite{boyd2004convex}} A function $f$ is called quasi-concave if
\[f(\lambda x+(1-\lambda )y)\geq \min {\big \{}f(x),f(y){\big \}}\]
Quasi-concave functions satisfy the following properties:
\begin{itemize}
    \item If $f$ is quasi-concave, then the superlevel sets, i.e., sets $S$
    of the form $S = \{ z \mid f(z) \geq \gamma \}$ for some $\gamma$, are convex.
    \item If $f_1, f_2, \ldots, f_n$ are both quasi-concave functions then the point-wise minimum of these, i.e., $f = \min_{1 \leq i\leq n} f_i$ is also quasi-concave.
\end{itemize}

\paragraph{Absolute Moments Of Gaussian Random Variable}: The $i^{th}$ absolute moments of a random variable $x$ is given as $\E[\abs{x^i}]$. For Gaussian random variable with variance $\sigma^2$, the absolute moments are $\E[\abs{x^i}] = \frac{\sigma^i2^{i/2}}{\sqrt{\pi}}\Gamma(\frac{i+1}{2})$, where $\Gamma$ is the Gamma function. Some properties of the Gamma function are: $\Gamma(i) = (i-1)!$, $\Gamma(\frac{i+1}{2}) = \frac{i-1}{2}\Gamma(\frac{i-1}{2})$ for all $i \in \mathbb{N}_{>0}$.

\paragraph{Subgaussian Random Varibale/Vector} A random variable $x \in \rr$ is said to be sub-Gaussian with parameter $\sigma^2$ if $\mathbb{E}[x] = 0$ and its moment generating function satisfies $\E[e^{sx}] \leq e^{\frac{\sigma^2s^2}{2}} $. This is denoted as $x \sim \sg{\sigma^2}$. Then a random vector $X$ is considered to be sub-Gaussian with parameter $\sigma^2$ if for all unit vectors $v$, $v^TX \sim \sg{\sigma^2}$. With slight abuse of notation we denote this as $X \sim \sg{\sigma^2}$.

Sub-Gaussian random variables satisfy the following properties: 
\begin{itemize}
    \item If $x_1, x_2, \ldots, x_N$ are independent sub-Gaussian random variables with parameter $k$, then $\frac{\sum_{i=1}^N x_i}{N} \sim \sg{\frac{k}{N}}$.
    \item If $x \sim \sg{k}$, then $\prob(x > t) \leq e^{-\frac{t^2}{2k}}$ and $\prob(x < -t) \leq e^{-\frac{t^2}{2k}}$
    \item If $x_1, x_2, \ldots, x_N$ are not necessarily independent sub-Gaussian random variables with parameter $k$, then $\prob(\max_{1 \leq i \leq N} x_i > t) \leq Ne^{-\frac{t^2}{2k}}$.
\end{itemize}

\paragraph{Generalized Neyman Pearson Lemma \cite{chernoff1952generalization}}
In order to solve the optimization problems for our framework we use the Generalized Neymann Pearson Lemma \cite{chernoff1952generalization}. Here, we give the lemma with a simplified short proof

\begin{Lemma}[Generalized Neymann Pearson Lemma]
    Let $f_0, f_1, \ldots , f_m$ be real-valued, $\mu-$integrable functions defined on a Euclidean space $X$ . Let $\psi_0$ be any function of the form
    \begin{equation}
        \psi_0(x) \begin{cases}
            =  1, &\textbf{if } f_0(x) < k_1f_1(x) + \ldots + k_mf_m(x) \\
            = \gamma(x), &\textbf{if } f_0(x) = k_1f_1(x) + \ldots + k_mf_m(x) \\
            = 0, & \textbf{if } f_0(x) > k_1f_1(x) + \ldots + k_mf_m(x)
        \end{cases}
    \end{equation}
    where $0 \leq \gamma(x) \leq 1$. Then $\psi_0$ minimizes $\int_X \psi_0 f_0 d\mu$ over all $\psi, 0 \leq \psi \leq 1$ such that for $i = 1, \ldots, m$
    $$ k_i\int_X \psi f_i d\mu \geq k_i\int_X \psi_0 f_i d\mu$$
\begin{proof}
    We start by observing that under the given definition of $\psi_0, \psi$ the following inequality holds
    \begin{align}
    \label{ineq}
       \int_{X} (\psi - \psi_0) \bigg(f_0 - \sum_{i=1}^m k_if_i\bigg) d\mu &\geq 0 
    \end{align}
    We can show by proving $\forall x, (\psi(x) - \psi_0(x)) \bigg(f_0(x) - \sum_{i=1}^m k_if_i(x)\bigg) \geq 0$. We show this by doing a case analysis:
    \begin{itemize}
        \item If $f_0(x) - \sum_{i=1}^m k_if_i(x)) > 0$ then $\psi_0 (x) = 0$. As $\psi(x) \geq 0$, $\psi(x) - \psi_0(x) \geq 0$ making $(\psi(x) - \psi_0(x)) \bigg(f_0(x) - \sum_{i=1}^m k_if_i(x)\bigg) \geq 0$.
        \item If $f_0(x) - \sum_{i=1}^m k_if_i(x)) < 0$ then $\psi_0 (x) = 1$. As $\psi(x) \leq 1$, $\psi(x) - \psi_0(x) \leq 0$ making $(\psi(x) - \psi_0(x)) \bigg(f_0(x) - \sum_{i=1}^m k_if_i(x)\bigg) \geq 0$.
        \item Finally if $f_0(x) - \sum_{i=1}^m k_if_i(x)) = 0$ then $(\psi(x) - \psi_0(x)) \bigg(f_0(x) - \sum_{i=1}^m k_if_i(x)\bigg) = 0 \geq 0$.
    \end{itemize}
    Using the inequality \ref{ineq}, we see that
    \begin{align*}
        \int_{X} \psi \bigg(f_0 - \sum_{i=1}^m k_if_i\bigg) d\mu &\geq \int_{X} \psi_0 \bigg(f_0 - \sum_{i=1}^m k_if_i\bigg) d\mu \\
        \forall~i\geq 1,~k_i \int_{X} \psi f_i d\mu \geq k_i \int_{X} \psi_0 f_i d\mu &
        \implies \int_{X} \psi f_0 d\mu \geq \int_{X} \psi_0f_0  d\mu
    \end{align*}
\end{proof}
\end{Lemma}

\clearpage

\section{Proofs for A General Framework for Randomized Smoothing}

\subsection{Regularity Properties of $\mathbf{\G}$}
\begin{Theorem}
   If $\forall \alpha \in \mathbb{N}^d$, $\int_{\rr^d}\abs{D^{\alpha}\mu(z)}dz$ exists and is finite, then $\G \subset \mathcal{C}^\infty$. Moreover if $g \in \G$ is given as $g = f \star \mu$ for some $f \in \F$, then $$\nabla^i g(x) = \int_{\rr^d} f(y) (-1)^i(\nabla^i \mu) (y-x)  dy$$
\begin{proof}
It follows from the definition that whenever the right-hand side exists, we have 
\begin{equation*}
    \begin{aligned}
    g(x) &= \int_{\rr^d} f(x + z) \mu(z) dz\\
    &= \int_{\rr^d} f(y)\mu(y-x) dy, \\
    \nabla^i_x g(x) &= \nabla^i_x \int_{\rr^d} f(y)\mu(y-x) dy \\
    &=  \int_{\rr^d} f(y)\nabla^i_x\mu(y-x) dy.
\end{aligned}
\end{equation*}

As we have $i > 1$, we get $\nabla^i_x (y-x) = 0$ and 
\begin{align*}
    \nabla^i_x g(x) &= \int_{\rr^d} f(y)(\nabla^i\mu)(y-x) (\nabla_x (y-x))^i dy \\
    &= \int_{\rr^d} f(y) (-1)^i (\nabla^i\mu)(y-x) dy.
\end{align*}
In order to show that the integral exists and is finite we show that integral converges for every element of the tensor. Our problem reduces to showing that for all $\alpha\in \mathbb{N}^d$ the integral $\int_{\rr^d} f(y) (-1)^i (D^\alpha\mu)(y-x) dy$ exists and is finite. Using integrability conditions we see this is equivalent to showing
\begin{align*}
    \int_{\rr^d} \abs{(D^\alpha\mu)(y-x)} dy < \infty \quad
    \iff \quad \int_{\rr^d} \abs{D^\alpha\mu(z)} dz < \infty.
\end{align*}
\end{proof}
\end{Theorem}

\begin{Lemma}
\label{lem: deriv}
    Let $\mu_0(z)$ denote the Gaussian distribution $\mathcal{N}(0, \sigma^2)$, then
    $\frac{d^j}{dz^j} \mu_0 (z) = q_j(z) \mu_0(z)$ for some $j^{th}$ degree polynomial $q_j$ with finite coefficient $a_{j,i}$ for $0 \leq i \leq j$. Moreover, $a_{j+1,i} = (i+1)a_{j,i+1} - \frac{1}{\sigma^2} a_{j, i-1}$
    \begin{proof}
        For the base case $j = 1$, we see that  $\frac{d}{dz} \mu_0 (z) = -\frac{z}{\sigma^2} \mu_0(z)$. Thus, we we have $q_1(z) = -\frac{z}{\sigma^2}$ which is a degree 1 polynomial.
        
        For the inductive step, we see that $\frac{d^{j+1}}{dz^{j+1}} \mu_0 (z) = \frac{d}{dz} \big(\frac{d^j}{dz^j} \mu_0 (z)\big) = \frac{d}{dz} q_j(z) \mu_0(z) = \big(\frac{d}{dz} q_j(z)\big)\mu_0(z) + q_j(z)(-\frac{z}{\sigma^2} \mu_0(z)) = \big(\frac{d}{dz} q_j(z) -\frac{z}{\sigma^2}q_j(z)\big) \mu_0(z) $. Thus, $q_{i+1}(z) = \big(\frac{d}{dz} q_j(z) -\frac{z}{\sigma^2} q_j(z)\big) $ which is clearly a polynomial of degree $(j+1)$ with coefficients $a_{j+1, i} = (i+1)a_{j,i+1} - \frac{1}{\sigma^2} a_{j, i-1}$.
    \end{proof}
\end{Lemma}

\begin{Corollary}
    When $\mu$ is given as the isotropic Gaussian distribution $\mathcal{N}(0, \sigma^2 \mathcal{I})$, then 
    $\G \subset \mathcal{C}^\infty$ and $\nabla^i g(x) = \int_{\rr^d} f(y) (-1)^i\nabla^i \mu (y-x)  dy$.
\begin{proof}
    Using the fact that for isotropic gaussian we can write $\mu(x) = \Pi_{i=1}^d \mu_0(x_i)$ (where $\mu_0$ is also a gaussian pdf), we have
\begin{align*}
    \int_{\rr^d}|D^{\alpha} \mu(z)|dz &= 
    \int_{\rr} \abs{D^{\alpha} \Pi_{i=1}^d \mu_0 (z_i)} dz\\
    &=  \int_{\rr} \Pi_{i=1}^d \abs{\frac{d^{\alpha_i}}{dz_i^{\alpha_i}} \mu_0 (z_i)} dz\\
    &= \Pi_{i=1}^d \int_{\rr} \abs{\frac{d^{\alpha_i}}{dz_i^{\alpha_i}} \mu_0 (z_i)} dz_i.
\end{align*}
As we know that the product of $d$ finite values is finite, we only need to show that for any value $i \in \mathbb{N}$,
$\int_{\rr} |\frac{d^i}{dz^i} \mu_0 (z)| dz$ is finite. Using Lemma \ref{lem: deriv}, we see that it suffices to show that for all $j \in \mathbb{N}$, $\int_{\rr} |x^j \mu_0 (z)| dz$ is finite. But this quantity is known as the absolute central moment of normal distribution and is given by $\sigma^j \sqrt{\frac{2^j}{\pi}} \Gamma\big(\frac{1 + j}{2}\big)$ which is finite.
\end{proof}
\end{Corollary}

\begin{Lemma}
\label{lem: interchange}
    For $\mu$ given by isotropic Gaussian distribution $\mathcal{N}(0, \sigma^2 \mathbf{I})$ and a function $f \in \F$, if $\norm{x - w}_\infty \leq R$ for some finite $R$, then
    \[\int_{\rr^d} f(y) \sum_{\alpha \in \mathbb{N}^d} D^\alpha \mu (y-w) \frac{(w-x)^\alpha}{\alpha !}  dy = \sum_{\alpha \in \mathbb{N}^d} \frac{(w-x)^\alpha}{\alpha !} \int_{\rr^d} f(y)  D^\alpha \mu (y-w)  dy\]
\begin{proof}
    As this can be regarded as a double integral where the sum is an integral over the counting measure, we can use Fubini's Theorem to reduce it to proving : 
    \[\sum_{\alpha \in \mathbb{N}^d} \abs{\frac{(w-x)^\alpha}{\alpha !}} \int_{\rr^d} \abs{f(y)  D^\alpha \mu (y-w) dy} < \infty\]
    As $f(y)$ only takes values between 0 and 1, we see that 
    \begin{align*}
        \sum_{\alpha \in \mathbb{N}^d} \abs{\frac{(w-x)^\alpha}{\alpha !}} \int_{\rr^d} \abs{ f(y)  D^\alpha \mu (y-w)} dy 
        &\leq \sum_{\alpha \in \mathbb{N}^d} \abs{\frac{(w-x)^\alpha}{\alpha !}} \int_{\rr^d} \abs{ D^\alpha \mu (z)} dz \\
        &= \sum_{\alpha \in \mathbb{N}^d} \prod_{j=1}^d \Bigg(\abs{\frac{(w_j-x_j)^{\alpha_j}}{{\alpha_j} !}} \int_{\rr} \abs{ \frac{d^{\alpha_j}}{dz^{\alpha_j}} \mu_0 (z)} dz \Bigg) \\
        &=  \prod_{j=1}^d \Bigg(\sum_{k \in \mathbb{N}} \abs{\frac{(w_j-x_j)^k}{k!}} \int_{\rr} \abs{\frac{d^k}{dz^k} \mu_0 (z)} dz \Bigg) \\
        &\leq \Bigg( \sum_{k \in \mathbb{N}} \frac{R^k}{k!} \int_{\rr} \abs{\frac{d^k}{dz^k} \mu_0 (z)} dz \Bigg)^d
    \end{align*}
    As $d$ is finite number, it is sufficient to show that the infinite sum converges. Using Lemma \ref{lem: deriv}, we see that $\frac{d^k}{dz^k} \mu_0(z) = q_k(z) \mu_0(z)$ for some $k^{th}$ degree polynomial $q_k$. Let $a_{j,i}$ be the co-efficient of $z^i$ in the polynomial $q_j$. 
    \[ \int_{\rr} \abs{\frac{d^k}{dz^k} \mu_0 (z)} dz = \int_{\rr} \abs{q_k(z) \mu_0 (z)} dz \leq \sum_{i=0}^k \int_{\rr} \abs{a_{ki} z^i \mu_0 (z)} dz = \sum_{i=0}^k \abs{a_{ki}} \frac{\sigma^i 2^{i/2} \Gamma(\frac{i+1}{2})}{\sqrt{\pi}}\]
    Using comparison condition, it is sufficient to show that the sum $\sum_{k = 0}^{\infty} \frac{R^k}{k!} \sum_{i=0}^k \abs{a_{k,i}} \frac{\sigma^i 2^{i/2} \Gamma(\frac{i+1}{2})}{\sqrt{\pi}}$ converges. Now, we prove the convergence using the ratio test. Using the fact $a_{j+1,i} = (i+1)a_{j, i+1} - \frac{1}{\sigma^2}a_{j, i-1}$, we see
    \begin{align*}
        \frac{\frac{R^{k+1}}{(k+1)!}\sum_{i=0}^{k+1} \abs{a_{k+1,i}} \frac{2^{i/2} \sigma^{i+1}\Gamma(\frac{i+1}{2})}{\sqrt{\pi}}}{\frac{R^k}{k!} \sum_{i=0}^k \abs{a_{k,i}} \frac{2^{i/2} \sigma^{i+1}\Gamma(\frac{i+1}{2})}{\sqrt{\pi}}} &= \frac{R}{k+1} \frac{\sum_{i=0}^{k+1} \abs{a_{k+1,i}} 2^{i/2} \sigma^{i+1} \Gamma(\frac{i+1}{2})}{\sum_{i=0}^k \abs{a_{k,i}} 2^{i/2} \sigma^{i+1}\Gamma(\frac{i+1}{2})} \\
        &\leq \frac{R}{k+1} \frac{\sum_{i=0}^{k+1} ((i+1)\abs{a_{k,i+1}} + \frac{1}{\sigma^2}\abs{a_{k,i-1}}) 2^{i/2} \sigma^{i+1}\Gamma(\frac{i+1}{2})}{\sum_{i=0}^k \abs{a_{k,i}} 2^{i/2} \sigma^{i+1}\Gamma(\frac{i+1}{2})} \\
        &= \frac{R}{k+1} \frac{\sum_{i=0}^{k} \abs{a_{k,i}} (i \sigma^i\Gamma(\frac{i}{2}) 2^{(i-1)/2} + \sigma^{i}\Gamma(\frac{i+2}{2}) 2^{(i+1)/2})}{\sum_{i=0}^k \abs{a_{k,i}} 2^{i/2} \sigma^{i+1}\Gamma(\frac{i+1}{2})} \\
        & = \frac{R}{k+1} \frac{\sum_{i=0}^{k} 2i \abs{a_{k,i}} 2^{(i-1)/2} \sigma^i \Gamma(\frac{i}{2}) }{\sum_{i=0}^k \abs{a_{k,i}} 2^{i/2} \sigma^{i+1}\Gamma(\frac{i+1}{2})} \\
        \text{Using (\cite{wendel1948note}), $\frac{\Gamma(\frac{i}{2})}{\Gamma(\frac{i+1}{2})}
        \leq \frac{\sqrt{\frac{i+1}{2}}}{\frac{i}{2}}$,} & \\
        &\leq \frac{R}{k+1} \frac{\sum_{i=0}^{k}  2\sqrt{i+1} \abs{a_{k,i}} 2^{i/2} \sigma^i \Gamma(\frac{i+1}{2})}{\sum_{i=0}^k \sigma \abs{a_{k,i}} 2^{i/2} \sigma^i \Gamma(\frac{i+1}{2})} \\
        &\leq \frac{R}{k+1} \frac{2\sqrt{k+1}}{\sigma} = \frac{2R}{\sigma\sqrt{k+1}}
    \end{align*}
    For any finite $R$, $\lim_{k \rightarrow \infty} \frac{2R}{\sigma\sqrt{k+1}} = 0 < 1$. So the series is convergent.
\end{proof}
\end{Lemma}
\begin{Theorem}
   When $\mu$ is the isotropic Gaussian distribution $\mathcal{N}(0, \sigma^2 \mathcal{I})$, then $\forall g \in \G$, $g$ is a real analytic function with infinite radius of convergence ,i.e., the Taylor series of $g$ around any point $w$ converges to the function $g$ everywhere.
\begin{proof} Let us take the Taylor expansion at a point $w$. In order to show that the Taylor expansion has an infinite radius of convergence, we consider any arbitrarily big value $R$ and show that if $\norm{x - w}_\infty \leq R$, then 
\begin{align*}
    g(x) &= \int_{\rr^d} f(x + z)\mu(z) dz \\
    &= \int_{\rr^d} f(y) \mu(y-x)  dy
\end{align*}
    Using the Taylor expansion of the gaussian PDF and the fact it's radius of convergence is infinite
\begin{align*}    
    g(x) &= \int_{\rr^d} f(y) \sum_{\alpha \in \mathbb{N}^d} D^\alpha \mu (y-w) \frac{((y-x) - (y-w))^\alpha}{\alpha !}  dy
\end{align*}    
Now using Lemma \ref{lem: interchange}, we get
\begin{align*}
    g(x) &= \sum_{\alpha \in \mathbb{N}^d} \frac{(w-x)^\alpha}{\alpha !} \int_{\rr^d} f(y)  D^\alpha \mu (y-w)  dy
\end{align*}
Finally, we use Corollary \ref{cor: exist}, to get
\[g(x) = \sum_{\alpha \in \mathbb{N}^d} \frac{(x-w)^\alpha}{\alpha!} D^\alpha g(w) \]
As for any arbitrarily large $R$, the Taylor series converges for any $x$ satisfying $\norm{x - w}_\infty \leq R$ we see that the radius of convergence is infinite. Clearly, this holds for all points $w \in \rr^d$ and all $g \in \G$.
\end{proof}
\end{Theorem}

\subsection{Certification For Randomized Smoothing Using First-Order Information}

We use the Generalized Neymann Pearson Lemma (\cite{chernoff1952generalization}) to solve the optimization problem \ref{eq: GenNP}. 
\setcounter{equation}{5}
\begin{Theorem}[Lower Bound of $\pr(z)$]
    For a base classifier $f \in \F$, if $g = f \star \mu$, $\mu$ is the isotropic Gaussian distribution $\mathcal{N}(0, \sigma^2 \mathcal{I}), y^{(0)} = g(x)$, $y^{(1)} = \nabla g(x)$, then for any unit vector $v$ and any positive value of $r$, $\pr(x + \sigma rv)$ can be lower bounded by solving the following set of equations:
    \begin{align}
       \int_{-\infty}^\infty \frac{1}{\sqrt{2\pi}}e^{-\frac{x^2}{2}} \Phi(c(x)) dx &= q  
    \end{align}
    \begin{subequations}
    \begin{minipage}{0.5\textwidth}
        \begin{align}
            \int_{-\infty}^\infty
            \frac{1}{\sqrt{2\pi}}e^{-\frac{x^2}{2}} \frac{1}{\sqrt{2\pi}}e^{-\frac{c(x)^2}{2}} dx = m_2
        \end{align}
    \end{minipage}
    \begin{minipage}{0.5\textwidth}
        \begin{align}
        \int_{-\infty}^\infty \frac{1}{\sqrt{2\pi}}xe^{-\frac{x^2}{2}} \Phi(c(x)) dx = m_1
        \end{align}
    \end{minipage}
    \end{subequations}
    with $q \leq y^{(0)}, m_1 \leq \sigma v^Ty^{(1)}, m_2 \leq \sigma \norm{y^{(1)} - v^Ty^{(1)}v}_2$, $c(x) := c_0 + c_1 x + c_2e^{rx}$, and $\Phi(z)$ being the CDF of the standard normal distribution. If the solution $(c_0, c_1, c_2)$ of above equations has $c_1 < 0$, then the lower bound of $\pr(x + \sigma rv)$ is instead given by solving Equations \eqref{eq: probbnd} to \eqref{eq: ybnd} with $c(x) := c_0 + c_2e^{rx}$.
\begin{proof}
    In order to solve the Equation \ref{eq: GenNP} under the local constraints $H_1^{x_0}(h) = y^{(0)} - h(x_0) = 0$ and  $H_2^{x_0}(h) = y^{(1)} - \nabla h(x_0) = 0$. Setting the measure to be $\mu_{\sigma} = \big( \frac{1} {\sqrt{2\pi\sigma^2}} \big)^{\frac{d}{2}} e^{-\frac{\norm{z}_2^2}{2\sigma^2}}$ and using the fact that $h \in \G$, we have $h = f' \star \mu$ for some $f \in \F$. Thus, the constraints can be expressed using the base classifier as given as
    \begin{align*}
        \int_{\rr^d} f'(x_0 + z) d\mu_\sigma &= y^{(0)} \\
        \int_{\rr^d} \frac{z}{\sigma^2} f'(x_0 + z) d\mu_\sigma &= y^{(1)}
    \end{align*}
    and the optimization problem is given as 
    $\min_{f' \in \F} \int_{\rr^d} e^{-\frac{R^2}{2\sigma^2}} e^{\frac{Rv^Tz}{\sigma^2}} f'(x_0 + z) d\mu_\sigma$.
    
    In order to make the math simpler we use the following basis transformation
    we rotate the basis such that we have $z_1$ along the $v$, $z_2$ along $y^{(1)} - v^Ty^{(1)}v$ and then we scale the basis by a factor of $\frac{1}{\sigma}$.
    
    The constraints can now be expressed using $\mu = \big( \frac{1} {\sqrt{2\pi}} \big)^{\frac{d}{2}} e^{-\frac{\norm{z}_2^2}{2}}$ given as
    \begin{align*}
        \int_{\rr^d} f'(x_0 + z) d\mu &= y^{(0)} \\
        \int_{\rr^d} \frac{z_1}{\sigma} f'(x_0 + z) d\mu &=  v^Ty^{(1)} \\
        \int_{\rr^d} \frac{z_2}{\sigma} f'(x_0 + z) d\mu &=  \norm{y^{(1)} - v^Ty^{(1)}v}_2 \\
        \int_{\rr^d} \frac{z_i}{\sigma} f'(x_0 + z) d\mu &= 0, \text{ if } i \geq 3
    \end{align*}
    Then defining $r = \frac{R}{\sigma}$, the optimization problem is given as 
    $$\min_{f' \in \F} \int_{\rr^d} e^{-\frac{r^2}{2}} e^{rz_1} f'(x_0 + z) d\mu = e^{-\frac{r^2}{2}}\min_{f' \in \F} \int_{\rr^d} e^{rz_1} f'(x_0 + z) d\mu  $$
    Using the Generalized Neymann Pearson Lemma, we see that the minima occurs for the function $f_0$ such that $f_0(x_0+z) = 1$ if $ e^{rz_1} \leq a^Tz + b$ and $0$ otherwise, for some $a \in \rr^d, b \in \rr$ such that the constraints are satisfied. We can use the constraints to solve for $a,b$ in order to get the value of the minimization. 
    \begin{Claim}
        For $i \geq 3$, we can show that we need $a_i = 0$.
    \begin{proof}
        Assume to the contrary $a_i > 0$, then 
    \begin{align*}
        \int_{\rr^d} \frac{z_i}{\sigma} f'(x_0 + z) d\mu &= 
        \int_{e^{rz_1} \leq a^Tz + b} \frac{z_i}{\sigma} f'(x_0 + z) d\mu \\
         &= \int_{z_i \geq \frac{e^{rz_1}  - (a_{-i}^Tz_{-i} + b)}{a_i}} \frac{z_i}{\sigma} d\mu > 0
    \end{align*}
    Consider the function $l(z_{-i}) := \frac{e^{rz_1}  - (a_{-i}^Tz_{-i} + b)}{a_i}$. We know that for the standard normal measure $\mu_0$ gaussian cdf, the value of $\int_{z_i \geq l} \frac{z_i}{\sigma} d\mu_0 \geq 0$ for any value of $l$ with the equality holding only if $l = -\infty$. So, we see that $\int_{z_i \geq l(z_{-i})} \frac{z_i}{\sigma} d\mu \geq 0$ for any function $l(z_{-i})$ with equality holding only if $l(z_{-i})$ is $-\infty$ almost everywhere does not hold. So, we get a contradiction. Similarly we can also show a contradiction for the case when $a_i < 0$. Thus, we have that for $i \geq 3$, $a_i = 0$.
    \end{proof}
    \end{Claim}
      Substituting the values of $a_i$ in our constraints and simplifying the integrals we have the following system of equations:
    \begin{align*}
        p_{x_0}(x_0 + \sigma rv) &= \int_{-\infty}^\infty \frac{1}{\sqrt{2\pi}}e^{-\frac{(x-r)^2}{2}} \Phi(c(x)) dx \\
        \int_{-\infty}^\infty \frac{1}{\sqrt{2\pi}}e^{-\frac{x^2}{2}} \Phi(c(x)) dx &= y^{(0)}\\
        \int_{-\infty}^\infty
        \frac{1}{\sqrt{2\pi}}e^{-\frac{x^2}{2}} \frac{1}{\sqrt{2\pi}}e^{-\frac{c(x)^2}{2}} dx &= \sigma \norm{y^{(1)} - v^Ty^{(1)}v}_2 \\
        \int_{-\infty}^\infty \frac{1}{\sqrt{2\pi}}xe^{-\frac{x^2}{2}} \Phi(c(x)) dx &= \sigma v^Ty^{(1)} 
    \end{align*}
    where $c(x) := \frac{b}{a_2} + \frac{a_1}{a_2} x + \frac{-1}{a_2}e^{rx}$ and $\Phi(z)$ denotes the CDF of the standard normal distribution.
    
    Although this gives a solution for $p_{x_0}(x_0 + \sigma rv)$ the constraints here have equalities which require us to get exact values of $y^{(0)}, y^{(1)}$. This is not possible to achieve in practice. In practice, we can only get a high confidence interval estimate of the values. So, we need to be able to solve for $p_{x_0}(x_0 + \sigma rv)$ given interval estimates of the parameters. 
    
    We notice that using the same argument in the Claim, we can use the constraint $\int_{-\infty}^\infty        \frac{1}{\sqrt{2\pi}}e^{-\frac{x^2}{2}} \frac{1}{\sqrt{2\pi}}e^{-\frac{c(x)^2}{2}} dx \geq 0$ to show that $a_2 > 0$. Similarly we have that if $y^{(0)} > 0$, then $b > 0$. Otherwise if $b < 0$, then we see that $c(x) < \frac{a_1}{a_2}x$ giving $\int_{-\infty}^\infty \frac{1}{\sqrt{2\pi}} e^{-\frac{x^2} {2}} \Phi(c(x)) dx < \int_{-\infty}^\infty \frac{1}{\sqrt{2\pi}}e^{-\frac{x^2}{2}} \Phi(\frac{a_1}{a_2}x) dx = 0.5$. As the coefficients $a_2, b > 0$, Generalized Neymann Pearson Lemma allows us to use lower bounds $p \leq y^{(0)}$ and $m_2 \leq \sigma \norm{y^{(1)} - v^Ty^{(1)} v}_2$ in the constraints to still get a valid estimate of $p_{x_0}(x_0 + \sigma rv)$. 
    
    The only variable that can be both negative and positive is $a_1$. If we use a lower bound $m_1 \leq \sigma v^Ty^{(1)}$ and the resulting solution has a positive value of $a_1$ then it is valid. However, we see that if we get a negative value of $a_1$ in the solution we can instead solve the relaxed minimization problem to get a lower bound of $p_{x_0}(x_0 + \sigma rv)$ without the constraint \ref{eq: xbnd}. We give this as
    \begin{align*}
        p_{x_0}(x_0 + \sigma rv) &= \int_{-\infty}^\infty \frac{1}{\sqrt{2\pi}}e^{-\frac{(x-r)^2}{2}} \Phi(c(x)) dx \\
        \int_{-\infty}^\infty \frac{1}{\sqrt{2\pi}}e^{-\frac{x^2}{2}} \Phi(c(x)) dx &= y^{(0)}\\
        \int_{-\infty}^\infty
        \frac{1}{\sqrt{2\pi}}e^{-\frac{x^2}{2}} \frac{1}{\sqrt{2\pi}}e^{-\frac{c(x)^2}{2}} dx &= \sigma \norm{y^{(1)} - v^Ty^{(1)}v}_2 \\
    \end{align*}
    where $c(x) := \frac{b}{a_2} + \frac{-1}{a_2}e^{rx}$.
\end{proof}
\end{Theorem}

\begin{Proposition}
    The certified safety region, $\mathbf{SR}_L(x)$, calculated using the zeroth and first-order local information is convex, i.e., if $x_1, x_2 \in \mathbf{SR}_L(x)$ then $\frac{x_1 + x_2}{2} \in \mathbf{SR}_L(x)$.
\begin{proof}
    \[ \mathbf{SR}_L(x) = \{ z \mid \pr(z) > 0.5\} \]
    So, $\mathbf{SR}_L(x)$ is a superlevel set of $\pr$. In order to show $\mathbf{SR}_L(x)$ is convex it is sufficient to show $\pr$ is a quasi-concave function. Using the definition of $\pr$, we have
    \[\pr\big( z \big) = \mathop{\boldsymbol\min}_{h \in \G}  h(z)_c \quad \textbf{ s.t.} \quad h(x) = y^{(0)}, \nabla h(x) = y^{(1)} \]
    \begin{Claim}
    For the lower bound probability function $\pr$ calculated using the zeroth and first-order information, $\pr(z) = (f \star \mu) (z)$ for some $f$ such that $f \star \mu$ satisfies all the optimization constraints and $f \star \mu $ is quasi-concave.
    \begin{proof}
    Using Generalized Neyman Pearson Lemma, we see that the minima of the constrained optimization problem occurs for some function $f$ that satisfies $f(x) = 1$ if $e^{\frac{2 z^Tx - \norm{z}_2^2}{2 \sigma^2}} \leq a^Tx + b$ and $0$ otherwise. Thus, $\pr(z) = (f \star \mu) (z)$ where $f$ is the indicator function for the set 
    \[S = \{ x \mid x \in \rr^d; e^{\frac{2 z^Tx - \norm{z}_2^2}{2 \sigma^2}} \leq a^Tx + b \} \]
    It is easy to see that the function $e^{\frac{2 z^Tx - \norm{z}_2^2}{2 \sigma^2}} - a^Tx - b$ is convex as the Hessian is given as $\frac{zz^T + \sigma^2 \mathbf{I}}{\sigma^4}e^{\frac{2 z^Tx - \norm{z}_2^2}{2 \sigma^2}} \succ 0 $. Thus, the set $S$ being a level set of a convex function is also convex. So, $f$ is the indicator function of a convex set and thus a log-concave function. Moreover, we have that $\mu$ being isotropic gaussian distribution is also log-concave. From the properties of log-concave functions we get that the convolution $f \star \mu$ is also log-concave and as log-concave functions are also quasi-concave, $f \star \mu$ is quasi-concave.
    \end{proof} 
    \end{Claim}
    
    Using this claim we see that as at every point $z$, $\pr(z) = (f \star \mu) (z)$ for some $f \star \mu$ that satisfies all the constraints and is also quasi-concave, we can add an extra constraint to get
    \[\pr\big( z \big) = \mathop{\boldsymbol\min}_{g \in \G}  g(z)_c \quad \textbf{ s.t.} \quad g(x) = y^{(0)}, \nabla g(x) = y^{(1)}, \quad g(x)_c \textbf{ is quasi-concave} \]
    
    to get the same $\pr$. As $\pr$ can be written as the minima over a set of quasi-concave functions, we see that by the property of quasi-concave functions, $\pr$ is also quasi-concave. Thus, we see that $\mathbf{SR}_L(x)$ is convex.
    \end{proof}
\end{Proposition}

\begin{Proposition}
    For any given value of $y^{(0)}, y^{(1)}$, the directional robustness along $v$, $\R{v}$, given by the first-order certification method is a non-increasing function of the angle between $v$ and $y^{(1)}$\mbox{, i.e., $\cos^{-1}\Big(\frac{v^Ty^{(1)}}{\norm{v}_2\norm{y^{(1)}}_2}\Big)$}.
\begin{proof}
    It follows from Theorem \ref{thm: direc} that given some value of $y^{(0)}$ and $y^{(1)}$ the minimal probability $p_{x_0}(x_0 + rv)$ at distance $r$ along a direction $v$ depends only on the angle between $v$ and $y^{(1)}$. Given some fixed $y^{(0)}, y^{(1)}$ we can write $p_{x_0}(x_0 + rv)$ as a function of $\theta, p_{x_0}(x_0 + rv) =p_x(r, \theta)$. Given this we claim
    \begin{Claim}
        For any given value of $r$, $\frac{\partial p(r, \theta)}{\partial \theta} \leq 0$.
    \begin{proof}
        As it is easier to state our theorems for vectors, We relate $\theta$ back to our vectors using a vector valued function $w(\alpha)$ that gives us vectors in some plane $P$ containing $y^{(1)}$ such that the angle between $y^{(1)}$ and $w(\alpha)$ is $\alpha$.
        Now given any angle $\alpha$ and some distance $r$, Theorem \ref{thm: direc} gives us that there exists some function $f_0$ such that for $g_0 = f_0 \star \mu$ all the local constraints are satisfied and $g_0(x_0 + r w(\alpha)) = p(r, \alpha)$.
        As $p(r, \theta)$ gives the minimum value that can be assigned $x_0 + rw(\theta)$ by a function satisfying the given constraints, we see the $p(r, \theta) \leq g_0(x_0 + rw(\theta))$. So we see that if
        $\frac{\partial g_0(x_0 + rw(\theta))}{\partial \theta} \bigm |_{\theta = \alpha} \leq 0$ then  $\frac{\partial p(r, \theta)}{\partial \theta} \bigm |_{\theta = \alpha}\leq 0$. It is sufficient to show that 
        $\frac{\partial g_0(x_0+ rw(\theta))}{\partial \theta}\bigm |_{\theta = \alpha} \leq 0$. In order to make the make the calculation simpler we can do the same basis transformation as in proof of Theorem \ref{thm: direc}. Under the new basis we have $\frac{\partial g_0(x_0 + rw(\theta))}{\partial \theta}\bigm |_{\theta = \alpha} = - \frac{\partial g_0(x_0 + (z_1, z_2))}{\partial z_2}\bigm |_{z = (r, 0)}$ where $(z_1, z_2)$ is a two dimensional vector $z_1$ along the old $w(0)$ and $z_2$ along $w(\frac{\pi}{2})$.
        
        For the new basis we see that the proof of Theorem \ref{thm: direc} also gives the form of $f_0$, i.e., there exist some constants $a_1, b \in \rr$ and $a_2 \in \rr^+$ such that $f_0(x_0 + z) = 1$ if $e^{rz_1} \leq a_1z_1 + a_2z_2 + b$ and $0$ otherwise. Using this form we have
        \begin{align*}
            \frac{\partial g_0(x_0 + (z_1, z_2))}{\partial z_2}\bigm |_{z = (r, 0)} &= \frac{\partial}{\partial z_2} \int \int_{e^{rx} \leq a_1x + a_2y + b} \frac{1}{2\pi}e^{-\frac{(x - z_1)^2 + (y-z_2)^2}{2}} dy~dx \bigm |_{z = (r, 0)} \\
            &= \int \int_{e^{rx} \leq a_1x + a_2y + b} \frac{- y}{2\pi}e^{-\frac{(x - r)^2 + (y)^2}{2}} dy~dx \\
            &= \int_{-\infty}^\infty \int_{\frac{e^{rx} - a_1 x - b}{a_2} \leq y} \frac{- y}{2\pi}e^{-\frac{(x - r)^2 + (y)^2}{2}} dy~dx \\
            &= \int_{-\infty}^\infty \frac{1}{2 \pi} e^{-\frac{(x - r)^2}{2}} - e^{-\frac{(e^{rx} a_1 x - b)^2}{2a_2^2}} dx \geq 0 \\
            \frac{\partial g_0(x_0 + rw(\theta))}{\partial \theta}\bigm |_{\theta = \alpha} &= - \frac{\partial g_0(x_0 + (z_1, z_2))}{\partial z_2}\bigm |_{z = (r, 0)} \leq 0 \\
            \implies & \frac{\partial p(r, \theta)}{\partial \theta}\bigm |_{\theta = \alpha} \leq 0
        \end{align*}
    \end{proof}
    \end{Claim}
    Using the claim we can show that $\mathbf{R_v}$ is a non-increasing function of the angle $\cos^{-1}\Big(\frac{v^Ty^{(1)}}{\norm{v}_2\norm{y^{(1)}}_2}\Big)$ as follows: 
    For angle $\alpha$ let $\mathbf{R_{w(\alpha)}} = r$, then $p_{x_0}(x_0 + rw(\alpha)) = 0.5$. Using the claim we see that for any value of $\beta > \alpha$, $p_{x_0}(x_0 + rw(\beta)) = p(r, \beta) \leq p(r, \alpha) = 0.5$. So we conclude that for any $\beta > \alpha$, $\mathbf{R_{w(\beta)}} \leq r = \mathbf{R_{w(\alpha)}}$.
\end{proof}
\end{Proposition}

\begin{Corollary}[Certified $\ell_2$ Norm Radius]
    For a base classifier $f \in \F$, if $g = f \star \mu$, where $\mu$ is the isotropic Gaussian distribution $\mathcal{N}(0, \sigma^2 \mathcal{I}), y^{(0)} = g(x)$, $y^{(1)} = \nabla g(x)$, the $\ell_2$ norm radius $R$ is given as $R = \sigma r$, where $(r, z_1, z_2)$ is the solution of the system of equations:
    \begin{equation}
        \Phi(z_1 - r) - \Phi(z_2 - r) = 0.5
    \end{equation}
    \begin{subequations}
    \begin{minipage}{0.5\textwidth}
        \begin{equation}
        \Phi(z_1) - \Phi(z_2) = q
        \end{equation}
    \end{minipage}
    \begin{minipage}{0.5\textwidth}
        \begin{equation}
        \frac{1}{\sqrt{2\pi}}e^{-\frac{z_2^2}{2}} -  \frac{1}{\sqrt{2\pi}}e^{-\frac{z_1^2}{2}} = m_1
        \end{equation}
    \end{minipage}
    \end{subequations}
    
    with $q \leq y^{(0)}$ and $m_1 \geq \sigma\norm{y^{(1)}}_2$. 
\begin{proof}
    Using Proposition \ref{thm: growth} we see that the minimum value of $\mathbf{R_v}$ occurs when $v^Ty^{(1)}$ is smallest. As $v^Ty^{(1)} \geq - \norm{v}_2\norm{y^{(1)}}_2$ with equality when $v = \frac{-y^{(1)}}{\norm{y^{(1)}}_2}$. Using Theorem \ref{thm: direc} to solve for $\mathbf{R_v}$ along this direction yields $m_2 = 0$. Using the same proof as in the first Claim in proof of Theorem 3 we have $a_2 = 0$. So we can rewrite $f_0(x) = 1$ if $e^{rx_1} \leq a_1x_1 + b$. Solving for $a_1, b$ under the constraints gives us the equations:
    \begin{align*}
        \int_{e^{rx_1} \leq a_1x_1 + b} \frac{1}{\sqrt{2\pi}} e^{-\frac{x^2}{2}} dx &= p \\
        \int_{e^{rx_1} \leq a_1x_1 + b} \frac{1}{\sqrt{2\pi}}xe^{-\frac{x^2}{2}} dx &= -m_1
    \end{align*}
    where As for all values of $a,b$ the solution to the equation $e^{rx_1} \leq a_1x_1 + b$ is an interval of the form $[z_2, z_1]$, we can re-write the constraints as
    \begin{align*}
        \Phi(z_1) - \Phi(z_2) &= p \\
        \frac{1}{\sqrt{2\pi}}(e^{-\frac{z_2^2}{2}} - e^{-\frac{z_1^2}{2}}) &= m_1
    \end{align*}
    Using the resulting $f_0$ the minimum value of $g_0$ at $r$ is given as $\Phi(z_1 - r)  - \Phi(z_2 - r)$ which can be equated to $0.5$ to give the radius.
\end{proof}
\end{Corollary} 

\begin{Corollary}[Certified $\ell_1$ Norm Radius]
    For a base classifier $f \in \F$, if $g = f \star \mu$, where $\mu$ is the isotropic Gaussian distribution $\mathcal{N}(0, \sigma^2 \mathcal{I}), y^{(0)} = g(x)$, $y^{(1)} = \nabla g(x)$, the $\ell_1$ norm radius $R$ is obtained by solving $\pr(x + Rv) = 0.5$, where $\pr(x + Rv)$ is given by solving the problem in Theorem \ref{thm: direc} with \mbox{$m_1 \leq -\sigma\norm{y^{(1)}}_\infty$}, $m_2 \leq \sigma \sqrt{\norm{y^{(1)}}^2_2 - \norm{y^{(1)}}_\infty^2}$.
\begin{proof}
    We see that if the minimum directional robustness  among the basis vectors is given as $R_{\min} = \min_i \big( \min(\R{e_i}, \R{-e_i}) \big)$, then along every basis vector direction $\R{v} \geq R_{\min}$. Thus, the points $\{R_{\min} e_i, -R_{\min} e_i \mid 1 \leq i \leq d \} \subset \mathbf{SR}_L$ and by Proposition \ref{thm: convex}, the convex hull of these points the $\ell_1$ norm ball of radius $R_{\min}$ is also in $\mathbf{SR}_L$. Thus, the $\ell_1$ norm certified radius can be given as $\min_i \big( \min(\R{e_i}, \R{-e_i}) \big)$.
    
    Using Proposition \ref{thm: growth}, we see that this minimum occurs in the direction with the largest angle with $y^{(1)}$. So, the projection of $y^{(1)}$ along this direction can be given as $\min_i \min(e_i^T y^{(1)}, - e_i^T y^{(1)}) = - \max_i \max(-e_i^Ty^{(1)}, e_i^Ty^{(1)}) = - \norm{y^{(1)}}_\infty$. Now, we can use Theorem \ref{thm: direc} to give us the final solution.
\end{proof}
\end{Corollary}

\begin{Corollary}[Certified $\ell_\infty$ Norm Radius]
    For a base classifier $f \in \F$, if $g = f \star \mu$, where $\mu$ is the isotropic Gaussian distribution $\mathcal{N}(0, \sigma^2 \mathcal{I}), y^{(0)} = g(x)$, $y^{(1)} = \nabla g(x)$, the $\ell_\infty$ norm radius $R$ is obtained by solving $\pr(x + Rv) = 0.5$, where $\pr(x + Rv)$ is given by solving the problem in Theorem \ref{thm: direc} with \mbox{$m_1 \leq -\frac{\sigma}{\sqrt{d}}\norm{y^{(1)}}_1, m_2 \leq \frac{\sigma}{\sqrt{d}} \sqrt{d\norm{y^{(1)}}^2_2 - \norm{y^{(1)}}_1^2}$}.
\begin{proof}
    Consider the set of vectors $S = \{ v \mid \abs{v_i} = \frac{1}{\sqrt{d}} \}$. We see that if the minimum directional robustness among the vectors in $S$ is given as $R_{\min} = \min_{v \in S} \R{v}$, then along every vector direction $v$ in $S$, $\R{v} \geq R_{\min}$. Thus, the points $\{R_{\min} v \mid v \in S\} \subset \mathbf{SR}_L$ and by Proposition \ref{thm: convex}, the convex hull of these points the $\ell_{\infty}$ norm ball of radius $R_{\min}$ is also in $\mathbf{SR}_L$. Thus, the $\ell_{\infty}$ norm certified radius can be given as $\min_{v \in S} \R{v}$.
    
    Using Proposition \ref{thm: growth}, we see that this minimum occurs in the direction with the largest angle with $y^{(1)}$. So, the projection of $y^{(1)}$ along this direction can be given as $\min_{v \in S} v^T y^{(1)} = - \max_{v \in S} (-v)^Ty^{(1)} = - \max_{v \in S} v^Ty^{(1)} = - \frac{1}{\sqrt{d}}\norm{y^{(1)}}_1$. Now, we can use Theorem \ref{thm: direc} to give us the final solution.
\end{proof}
\end{Corollary}

\begin{Corollary}[Subspace Certified $\ell_p$ norm radius]
    For a base classifier $f \in \F$, if $g = f \star \mu$, where $\mu$ is the isotropic Gaussian distribution $\mathcal{N}(0, \sigma^2 \mathcal{I}), y^{(0)} = g(x)$, $y^{(1)} = \nabla g(x)$, and a subspace $S$ with orthogonal projection matrix $P_S$, for $p= 1, 2, \infty$ the subspace $\ell_p$ norm certified radius $R$ is obtained by solving $\pr(x + Rv) = 0.5$, where $\pr(x + Rv)$ is given by solving the problem in Theorem \ref{thm: direc} with $m_1 \leq -\sigma\norm{P_Sy^{(1)}}_{p'}, m_2 \leq \sigma \sqrt{\norm{y^{(1)}}_2^2 - \norm{P_Sy^{(1)}}_{p'}^2}$ and $\|\cdot\|_{p'}$ is the dual norm of $\|\cdot\|_{p}$.
\begin{proof}
    For orthogonal projection $P_S$ onto a subspace $S$, we can consider the vector $P_S y^{(1)}$ instead of $y^{(1)}$, the using almost identical arguments as before we get the corresponding projections and we can solve for the certified radii using Theorem \ref{thm: direc}.
\end{proof}
\end{Corollary}

\clearpage
\section{Theoretical Case Study:  Binary Linear Classifier}
\label{sec: Case_study}
\begin{figure}[h!]
    \centering
    \includegraphics[width=0.48\textwidth]{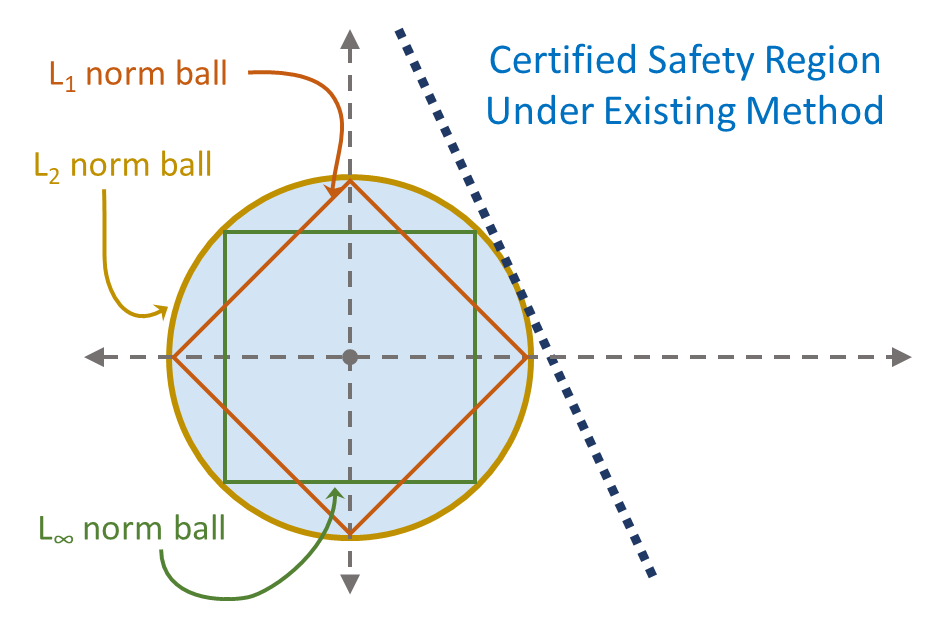}
    \includegraphics[width=0.48\textwidth]{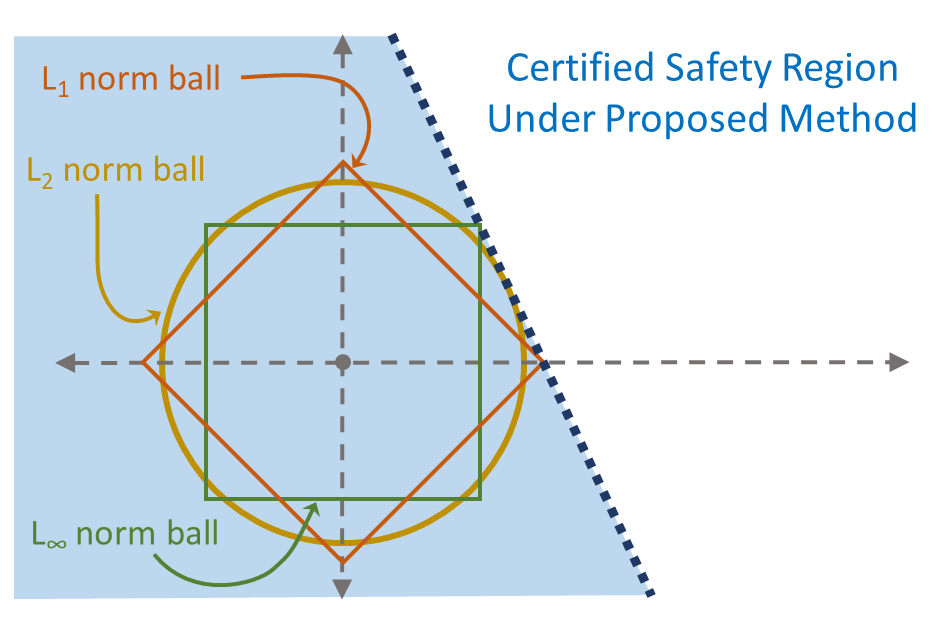}
    \caption{Certified safety regions for binary linear classifiers (input point $x$ grey circle at origin)}
    \label{fig: Compare}
\end{figure}

Given any binary linear classifier, $f(x) = \mathbf{1}_{w^Tx + b \leq 0}$ let $g$ defined by $f \star \mu$ where $\mu$ is the isotropic Gaussian distribution $\mathcal{N}(0, \sigma^2 \mathbf{I})$ (for any $\sigma$) be the smoothed classifier. For this case, Cohen et al. \cite[Appendix B]{Cohen2019Certified} showed the following:
\begin{itemize}
    \item The prediction of $g$ is same as the prediction of $f$, i.e., $\forall ~x \in \rr^d, ~f(x) = 1 \Leftrightarrow g(x)>0.5$.
    \item The $\ell_2$ norm certified radius for $g$ at $x_0$ is given as $R = \frac{\abs{w^Tx_0 + b}}{\norm{w}_2}$.
\end{itemize}

We saw in subsection \ref{subsec: firstRS}, the certified safety region calculated using only the zeroth order information has spherical symmetry. Thus, the certified safety region for $g$ at $x_0$ calculated using existing methods is a sphere of radius $R = \frac{\abs{w^Tx_0 + b}}{\norm{w}_2}$ centered at $x$. For the proposed method we show using both the zeroth and the first order information gives :

\begin{Proposition}
\label{prop: optimal}
    Under our proposed method, the certified safety region for $g$ at a point $x_0$ is given as the halfspace $H = \{x \mid sign(w^Tx + b) = sign(w^Tx_0 + b) \}$.
\begin{proof}
    In this case, we can calculate 
    \[y^{(0)} = \Phi\Bigg(\frac{\abs{w^Tx_0 + b}}{\sigma\norm{w}_2}\Bigg)\] 
    \[y^{(1)} = \frac{1}{\sqrt{2\pi\sigma^{2}}}e^{-\frac{(w^Tx_0 + b)^2}{2\sigma^2\norm{w}^2_2}} \frac{sign(w^Tx_0 + b)w}{\norm{w}_2}\]
    We shift the origin to $x_0$ and rotate and scale the basis by $\frac{1}{\sigma}$ to get a basis with positive $x_1$ along $sign(w^Tx_0 + b)w$. Then, we can use the framework and to calculate the feasible set of $g'$'s. Any valid $g'$ can be written as $g' = f' \star \mu$ where
    \begin{align*}
        \int_{\rr^d} f'(x) \Big(\frac{1}{2\pi\sigma^2}\Big)^{d/2} e^{-\frac{\norm{x}_2^2}{2}} dx &= p = y^{(0)} \\
        \int_{\rr^d} f'(x) \Big(\frac{1}{2\pi\sigma^2}\Big)^{d/2} x_1e^{-\frac{\norm{x}_2^2}{2}} dx &= m = \sigma \norm{y^{(1)}}_2 = \frac{1}{\sqrt{2\pi}}e^{-\frac{(\Phi^{-1}(1 - p))^2}{2}}
    \end{align*}
    Let $c = \Phi^{-1}(1 - p)$ and $f_0 = \mathbf{1}_{x_1 > c}$. It is easy to check that $f_0$ satisfies the above-mentioned constraints. We show that any $f'$ that satisfies the two constraints equals to $f$ almost everywhere. This is equivalent to saying $f_0 - f'$ is $0$ almost everywhere. We see
    \begin{align*}
        \int_{\rr^d} (f'(x) - f_0(x)) \Big(\frac{1}{2\pi\sigma^2}\Big)^{d/2} e^{-\frac{\norm{x}_2^2}{2}} dx &= 0 \\
        \int_{\rr^d} (f'(x) - f_0(x)) \Big(\frac{1}{2\pi\sigma^2}\Big)^{d/2} x_1e^{-\frac{\norm{x}_2^2}{2}} dx &= 0
    \end{align*}
    Moreover, we have that for $x_1 > c$, $f'(x) - f_0(x) \leq 0$ and for $x_1 \leq c$, $f'(x) - f_0(x) \geq 0$. Thus, we can rewrite the first constraint as
    \[ \int_{x_1 > c} \abs{f'(x) - f_0(x)} d\mu(x) = \int_{x_1 \leq c} \abs{f'(x) - f_0(x)} d\mu(x) \]
    For brevity we replaced the Gaussian integral over Lebesgue measure with an integral over the Gaussian measure. Now, for the second constraint we can re-write it as 
    \begin{align*}
        \int_{\rr^d} (f'(x) - f_0(x))x_1 d\mu(x) &= \int_{x_1 \leq c} \abs{f'(x) - f_0(x)}x_1 d\mu(x) - \int_{x_1 > c} \abs{f'(x) - f_0(x)}x_1 d\mu(x) \\
        &\leq c \int_{x_1 \leq c} \abs{f'(x) - f_0(x)} d\mu(x) - c \int_{x_1 > c} \abs{f'(x) - f_0(x)} d\mu(x) \\
        &= c (\int_{x_1 \leq c} \abs{f'(x) - f_0(x)} d\mu(x) - \int_{x_1 > c} \abs{f'(x) - f_0(x)} d\mu(x)) \\
        &= 0 \\
        \implies 0 &= \int_{\rr^d} (f'(x) - f_0(x))x_1 d\mu(x) \leq 0
    \end{align*}
    Thus the equality must hold in all the equations. Thus $\int_{x_1 > c} \abs{f'(x) - f_0(x)}x_1 d\mu(x) = c \int_{x_1 > c} \abs{f'(x) - f_0(x)} d\mu(x) $ which means $\int_{x_1 > c} \abs{f'(x) - f_0(x)} d\mu(x) = 0$. Then using the results from the first constraint $\int_{x_1 \leq c} \abs{f'(x) - f_0(x)} d\mu(x) = 0$. Thus, 
    \[\int_{\rr^d} \abs{f'(x) - f_0(x)} d\mu(x) = 0 \]
    As a result, $f'$ is equal to $f(0)$ almost everywhere w.r.t the Gaussian measure $\mu$. Thus, $g' = f' \star \mu = f_0 \star \mu $. Thus, we have only one feasible solution for $g'$ which is $g$. Thus, forall $x \in \rr^d$ $\pr(x) = g(x)$ and the certified safety region 
    \[\mathbf{SR}(x_0) = \{x \mid x \in \rr^d; ~ \pr(x) > 0.5\} = \{x \mid x \in \rr^d; ~ g(x) > 0.5\} \]
    Finally using the form of $g$ from \cite[Appendix B]{Cohen2019Certified}
    \[\mathbf{SR}(x_0) = \{x \mid sign(w^Tx + b) = sign(w^Tx_0 + b) \}\]
\end{proof}
\end{Proposition}

\subsection{Discussion}

Using the result from Proposition \ref{prop: optimal} we see that using both zeroth and first order information allows us to give the optimal certified safety region for binary linear classifiers. 

Although the results from zeroth order information give us the optimal $\ell_2$ radius as seen in Figure \ref{fig: Compare}, the radius for other threat models like $\ell_1, \ell_\infty$ can be sub-optimal. Using additional first-order information allows us to overcome this problem. As seen in Figure \ref{fig: Compare}, the safety region we achieve using the proposed work provides optimal radius for all $\ell_1, \ell_2, \ell_\infty$ threat models.

\clearpage
\section{Proofs for Numerical Estimation of First-Order Information}

\begin{Theorem}
   Given a black-box classifier $f$ and the random vector $z = w(f(x+w)_c - \frac{1}{2}) $ where $w \sim \mathcal{N}(0, \sigma^2 \mathbf{I})$, we have that $z - \sigma^2y^{(1)}$ is a sub-gaussian random vector with parameter $k = \sigma^2(\frac{1}{4} + \frac{3}{\sqrt{8\pi e}})$. For convenience, we do some abuse of notation to denote this as $(z - \sigma^2y^{(1)}) \sim \sg{k}$.
\begin{proof}
    For any unit norm vector $v$ consider the moment generating function for the variable
    \begin{align*}
        \E[e^{sv^T(z- \sigma^2y^{(1)})}] &= e^{-sv^T\sigma^2y^{(1)}}
        \E[e^{sv^Tz}]
    \end{align*}
    As the black-box classifier has binary output for every class, i.e, $f(x+w)_c$ is either 0 or 1, we have $e^{sv^Tz} = (e^{\frac{sv^Tw}{2}} - e^{\frac{-sv^Tw}{2}})f(x+w)_c + e^\frac{-sv^Tw}{2}$
    \begin{align*}
         \E[e^{sv^T(z- \sigma^2y^{(1)})}]   &= e^{-s\sigma^2v^Ty^{(1)}} \E[(e^{\frac{sv^Tw}{2}} - e^{\frac{-sv^Tw}{2}})f(x+w) + e^\frac{-sv^Tw}{2}] \\
            &= e^{\frac{s^2\sigma^2}{8} - s\sigma^2v^Ty^{(1)}} (1 + e^{\frac{-s^2\sigma^2}{8}}\E[(e^{\frac{sv^Tw}{2}} - e^{\frac{-sv^Tw}{2}})f(x+w)])
    \end{align*}
    Using Generalized Neymann-Pearson Lemma with the condition $\E[v^Tw f(x+w)] = \sigma^2v^Ty^{(1)} $, we have 
    $$ e^{-\frac{s^2\sigma^2}{8}}\E[(e^{\frac{sv^Tw}{2}} - e^{\frac{-sv^Tw}{2}})f(x+w)] \leq \frac{1}{\sqrt{2\pi\sigma^2}} \int_{\frac{-\sigma^2\abs{s}}{2}}^{\frac{\sigma^2\abs{s}}{2}} e^{-\frac{(-\beta + z)^2}{2\sigma^2}} + e^{-\frac{(\beta + z)^2}{2\sigma^2}} - e^{-\frac{z^2}{2\sigma^2}} dz  $$
    where $s\sigma^2v^Ty^{(1)} = \frac{\sigma^2\abs{s}}{\sqrt{2\pi\sigma^2}}(e^{-\frac{(-\beta)^2}{2\sigma^2}} + e^{-\frac{\beta^2}{2\sigma^2}} - e^{-\frac{0^2}{2\sigma^2}}) $.
    
    Let $\phi(w) = \frac{1}{\sqrt{2\pi\sigma^2}} e^{-\frac{(-\beta + w)^2}{2\sigma^2}} + e^{-\frac{(\beta + w)^2}{2\sigma^2}} - e^{-\frac{w^2}{2\sigma^2}}$. Then we have
    
    \begin{align*}
        \E[e^{sv^T(z - \sigma^2y^{(1)})}] &= e^{\frac{s^2\sigma^2}{8} - sv^T\sigma^2y^{(1)}} (1 + e^{\frac{-s^2\sigma^2}{8}}\E[(e^{\frac{sv^Tw}{2}} - e^{\frac{-sv^Tw}{2}})f(x+w)]) \\
        &\leq e^{\frac{s^2\sigma^2}{8} - \sigma^2\abs{s}\phi(0)} \bigg(1 + \int_{\frac{-\sigma^2\abs{s}}{2}}^{\frac{\sigma^2\abs{s}}{2}} \phi(w) dw\bigg) \\
        &\leq e^{\frac{s^2\sigma^2}{8}} e^{\int_{\frac{-\sigma^2\abs{s}}{2}}^{\frac{\sigma^2\abs{s}}{2}} (\phi(w) - \phi(0))dw}
    \end{align*}
    We see that the global Lipschitz constant for $\phi(w)$ is given as $\sup \norm{\phi'(w)} \leq 3 \sup \norm{\frac{w}{\sigma^2}\frac{1}{\sqrt{2\pi\sigma^2}}e^{\frac{-w^2}{2\sigma^2}}} = \frac{3}{\sigma^2 \sqrt{2\pi e}}$. Then we see that $\int_{\frac{-\sigma^2\abs{s}}{2}}^{\frac{\sigma^2\abs{s}}{2}} (\phi(w) - \phi(0))dw \leq \frac{3}{\sigma^2 \sqrt{2 \pi e}} \int_{\frac{-\sigma^2\abs{s}}{2}}^{\frac{\sigma^2\abs{s}}{2}} \abs{w}dw = \frac{3s^2\sigma^2}{4 \sqrt{2 \pi e}}$. Thus.
    \begin{align*}
         \E[e^{sv^T(z - \sigma^2y^{(1)})}] &\leq e^{\frac{s^2\sigma^2}{8}} e^{ \frac{3s^2}{4 \sqrt{2 \pi e}}} \\
         &= e^{ \frac{s^2}{2} \sigma^2(\frac{1}{4} + \frac{3}{\sqrt{8\pi e}})}
    \end{align*}
\end{proof}
\end{Theorem} 

\begin{Corollary}
    For any $\alpha$, let $Z_n$ be the empirical mean of $n$ samples of the random variable $z$, then given
    $t_1 = \sqrt{\frac{2kd(d\log{2} - \log{\alpha})}{n}}$,  $t_\infty = \sqrt{\frac{2k(\log{2d} - \log{\alpha})}{n}} $
    \begin{equation*}
        \begin{aligned}
            \prob \bigg(\abs{\norm{y^{(1)}}_1 - \norm{Z_n}_1} \leq t_1 \bigg) \geq 1 - \alpha, \quad  \prob \bigg(\abs{\norm{y^{(1)}}_\infty - \norm{Z_n}_\infty} \leq t_2 \bigg) \geq 1 - \alpha
        \end{aligned}
    \end{equation*}
\begin{proof}
    Using Theorem \ref{thm: subgauss} and the properties of subgaussian random vectors, we see that
    $Z_{n} \sim \sg{\frac{k}{n}}$. Let the set of vectors $S = \{ v \mid v \in \rr^d; \abs{v_i} = 1\}$,then $\norm{x}_1  = \max_{v \in S} v^Tx$. Using the maximal property of sub-gaussian random variables over the set of $2^d$ variables $\{ v^t Z_{n} \mid v \in S \}$ we get,
    \[\prob \bigg(\norm{y^{(1)} - Z_n}_1 \leq t_1 \bigg) \geq 1 - \alpha\]
    By the triangle inequality $\abs{\norm{y^{(1)}}_1 - \norm{Z_n}_1} \leq \norm{y^{(1)} - Z_n}_1$ we get the first result. Similarly, $\norm{x}_\infty  = \max_{i} \max(e_i^Tx, -e_i^Tx)$ and once again using the maximal property of sub-gaussian random variables over the set of $2d$ variables $\{ e_i^T Z_n, -e_i^TZ_n \mid e_i \text{ is a basis vector} \}$ we get, 
    \[\prob \bigg(\norm{y^{(1)} - Z_n}_\infty \leq t_\infty \bigg) \geq 1 - \alpha\]
    Again using triangle inequality, we see $\abs{\norm{y^{(1)}}_\infty - \norm{Z_n}_\infty} \leq \norm{y^{(1)} - Z_n}_\infty$ proving the second inequality.
    
\end{proof}
\end{Corollary}

\begin{Lemma}
\label{lem: bound}
    If we have two sub-gaussian random vectors $X \sim \sg{k_1}, Y \sim \sg{k_2}$ then
    \begin{align*}
        \prob(X^TY < -t) \leq \max\bigg(e^{-\frac{t^2}{\sqrt{2}dk_1k_2}}, e^{-\frac{t}{4\sqrt{k_1k_2}}}\bigg), \quad
        \prob(X^TY > t) \leq \max\bigg(e^{-\frac{t^2}{\sqrt{2}dk_1k_2}}, e^{-\frac{t}{4\sqrt{k_1k_2}}}\bigg)
    \end{align*}
    
\begin{proof}
    Consider the moment generating function for the variable $X^TY$. We have for $\abs{s} \leq \sqrt{\frac{2}{k_1k_2}}$
    \begin{align*}
        \E[e^{sX^TY}] &\leq
        \E[e^{\frac{k_2s^2\norm{X}^2}{2}}] \\
        &= \E[\E_{a \sim \mathcal{N}(0, r)}[e^{a^TX}]], \quad r = \frac{k_2s^2}{2} \\
        &= \E_{a \sim \mathcal{N}(0, r)}[\E[e^{a^TX}]]
        \leq \E_{a \sim \mathcal{N}(0, r)}[e^{\frac{k_1\norm{a}^2}{2}}] \\
        &= \Bigg(\frac{\frac{1}{\frac{1}{r} - k_1}}{r}\Bigg)^{\frac{d}{2}} 
        = \bigg(1 - \frac{s^2k_1k_2}{2}\bigg)^{\frac{-d}{2}}
    \end{align*}
    Now we see that
    \begin{align*}
        \prob(X^TY < -t) &= \prob(e^{-sX^TY} < e^{st}), s \geq 0 \\ 
        &\leq \frac{\E[e^{-sX^TY}]}{e^{st}} \\
        &\leq \bigg(1 - \frac{s^2k_1k_2}{2}\bigg)^{\frac{-d}{2}} e^{-st}
    \end{align*}
    Taking $s = \sqrt{\frac{d^2}{4t^2} + \frac{2}{k_1k_2}} - \frac{d}{2t}$ we see
    \begin{align*}
        \prob(X^TY < -t) &\leq \Bigg(1 + \frac{\sqrt{\frac{d^2}{4t^2} + \frac{2}{k_1k_2}} - \frac{d}{2t}}{2\frac{d}{2t}} \Bigg)^{\frac{d}{2}} e^{-\big(\sqrt{\frac{d^2}{4t^2} + \frac{2}{k_1k_2}} - \frac{d}{2t}\big)t} \\
        &\leq e^{\frac{d}{2}\frac{\sqrt{\frac{d^2}{4t^2} + \frac{2}{k_1k_2}} - \frac{d}{2t}}{2\frac{d}{2t}}} e^{-\big(\sqrt{\frac{d^2}{4t^2} + \frac{2}{k_1k_2}} - \frac{d}{2t}\big)t}\\
        &= e^{-\big(\sqrt{\frac{d^2}{4t^2} + \frac{2}{k_1k_2}} - \frac{d}{2t}\big)\frac{t}{2}} = e^{-\big(\sqrt{\frac{d^2k_1k_2}{8t^2} + 1} - \frac{d\sqrt{k_1k_2}}{\sqrt{8}t}\big) \frac{t}{\sqrt{2k_1k_2}}}\\
        &\leq e^{-\frac{t}{\sqrt{2k_1k_2}}\min\big(\frac{t}{d\sqrt{k_1k_2}}, \frac{1}{\sqrt{8}}\big)} \\
        \prob(X^TY < -t) &\leq \max\bigg(e^{-\frac{t^2}{\sqrt{2}dk_1k_2} }, e^{-\frac{t}{4\sqrt{k_1k_2}}}\bigg)
    \end{align*}
    We can use a similar proof to show 
    $$ \prob(X^TY > t)  \leq \max\bigg(e^{-\frac{t^2}{\sqrt{2}dk_1k_2} }, e^{-\frac{t}{4\sqrt{k_1k_2}}}\bigg)$$
\end{proof}
\end{Lemma}

\begin{Theorem}
    For any $\alpha \geq 2e^{-\frac{d}{16}}$, if we have two random vectors $X, Y$ such that $(X - \beta) \sim \sg{k_1}$ and $(Y - \beta) \sim \sg{k_2}$ then we can show that using
    $t = \sqrt{-\sqrt{2}k_1k_2d\log{\frac{\alpha}{2}}} $, $\epsilon_u = \sqrt{\frac{-(k_1 + k_2)\log{\frac{\alpha}{2}}}{2(X^TY + t)}}$, $\epsilon_l = \sqrt{\frac{-(k_1 + k_2)\log{\frac{\alpha}{2}}}{2(X^TY - t)}}$,
    \begin{equation*}
        \begin{aligned}
            \prob \bigg(\norm{\beta}_2 \leq \frac{\sqrt{X^TY + t}}{\sqrt{1 + \epsilon_u^2} - \epsilon_u}\bigg) \geq 1 - \alpha, \quad  \prob \bigg(\norm{\beta}_2 \geq \frac{\sqrt{X^TY - t}}{\sqrt{1 + \epsilon_l^2} + \epsilon_l}\bigg) \geq 1 - \alpha
        \end{aligned}
    \end{equation*}
\begin{proof}
    Using Lemma \ref{lem: bound}, we see $\prob \big(X^TY - \norm{\beta}_2^2 \leq -t + \beta\cdot((X - \beta) + (Y - \beta)) \big) \leq \max\bigg(e^{-\frac{t^2}{\sqrt{2}k_1k_2d}}, e^{-\frac{t}{4\sqrt{k_1k_2}}}\bigg)$. Taking $t = \sqrt{-\sqrt{2}k_1k_2d\log{\frac{\alpha}{2}}}$, we see that for $\alpha \geq 2e^{-\frac{d}{16}}$ the first term is bigger. So
    \begin{align*}
        \prob \bigg( \norm{\beta}_2^2 \leq X^TY + t - \beta\cdot(X + Y - 2\beta)) \geq 1 - \frac{\alpha}{2}
    \end{align*}
    From the sub-gaussian property of $X, Y$, we have
    $\prob (\beta\cdot(X + Y - 2\beta) \leq -t_1) \leq e^{-\frac{t_1^2}{2(k_1 + k_2) \norm{\beta}_2^2}}$. Taking $\epsilon_u = \sqrt{\frac{-(k_1+k_2) \log{\frac{\alpha}{2}}}{2(X^TY + t)}}$ and $t_1 = 2\epsilon_u\norm{\beta}_2\sqrt{X^TY}$, we get that 
    $$\prob \bigg( X^TY + t - \beta\cdot(X + Y - 2\beta) \leq 
    X^TY + t + 2\epsilon_u\norm{\beta}_2\sqrt{X^TY + t} \bigg) \geq 1 - \frac{\alpha}{2}$$
    
    Taking a union bound and combining the two inequalities we get
    \begin{equation}
        \begin{aligned}
        &\prob \bigg( \norm{\beta}_2^2 \leq X^TY + t + 2\epsilon_u\norm{\beta}_2\sqrt{X^TY + t} \bigg) &&\geq 1 - \alpha \\
        \iff & \prob \bigg( (1 + \epsilon_u^2)\norm{\beta}_2^2 \leq \Big(\sqrt{X^TY + t} +\epsilon_u\norm{\beta}_2\Big)^2 \bigg) &&\geq 1 - \alpha \\
        \iff & \prob \bigg( \norm{\beta}_2 \leq \frac{\sqrt{X^TY + t}}{\sqrt{1 + \epsilon_u^2} - \epsilon_u} \bigg) &&\geq 1 - \alpha
        \end{aligned}
    \end{equation}
    Using a similar proof we can also show that for $\epsilon_l = \sqrt{\frac{-(k_1 + k_2)\log{\frac{\alpha}{2}}}{2(X^TY - t)}}$, we have
    \begin{align*}
        \prob \bigg(\norm{\beta}_2 \geq \frac{\sqrt{X^TY - t}}{\sqrt{1 + \epsilon_l^2} + \epsilon_l}\bigg) \geq 1 - \alpha
    \end{align*}
\end{proof}
\end{Theorem}
Let $X = X_{n_1}, Y = Y_{n_2}$ be the empirical average of $n_1, n_2$ independent samples of the random variable $z$.

\begin{Corollary}
    For any $\alpha \geq 2e^{-\frac{d}{16}}$, given
    $t = \sqrt{-k^2\frac{\sqrt{2}d}{n_1n_2}\log{\frac{\alpha}{2}}} $, $\epsilon_u = \sqrt{\frac{-k(n_1 + n_2)\log{\frac{\alpha}{2}}}{2n_1n_2(X_{n_1}^TY_{n_2} + t)}}$, $\epsilon_l = \sqrt{\frac{-k(n_1 + n_2)\log{\frac{\alpha}{2}}}{2n_1n_2(X_{n_1}^TY_{n_2} - t)}}$,
    \begin{equation*}
        \begin{aligned}
            \prob \bigg(\norm{y^{(1)}}_2 \leq \frac{\sqrt{X_{n_1}^TY_{n_2} + t}}{\sqrt{1 + \epsilon_u^2} - \epsilon_u}\bigg) \geq 1 - \alpha, \quad  \prob \bigg(\norm{y^{(1)}}_2 \geq \frac{\sqrt{X_{n_1}^TY_{n_2} - t}}{\sqrt{1 + \epsilon_l^2} + \epsilon_l}\bigg) \geq 1 - \alpha
        \end{aligned}
    \end{equation*}
\begin{proof}
    Using Theorem \ref{thm: subgauss} and the properties of subgaussian random vectors, we see that
    $X_{n_1} \sim \sg{\frac{k}{n_1}}, Y_{n_2} \sim \sg{\frac{k}{n_2}}$. Then, using Theorem \ref{thm: estimator} we get the required values of $t, \epsilon_u, \epsilon_l$.
\end{proof}
\end{Corollary}

\clearpage
\section{Additional Experiments}

Here, we give additional experiments on the Imagenet dataset. We reuse the models given by Cohen et al. \cite{Cohen2019Certified} and calculate the certified accuracy at radius $R$ by counting the samples of the test set that are correctly classified by the smoothed classifier $g$ with certified radii of at least $R$. For both our proposed certificate and the baseline certificate~\cite{Cohen2019Certified}, we use a failure probability of $\alpha = 0.001$ and $N = 200,000$ samples for CIFAR and $N=1,250,000$ samples for Imagenet.

In the following plots we give a more detailed account of the improvement seen by using both the first and zeroth order information. For every trained model (depending on variance $\sigma$ used during training), we give the certified accuracy under $\ell_2$ norm threat model, $\ell_1$ norm threat model and the subspace $\ell_2$ norm threat model.

\begin{figure}[h!]
    \centering
    \includegraphics[width=\textwidth]{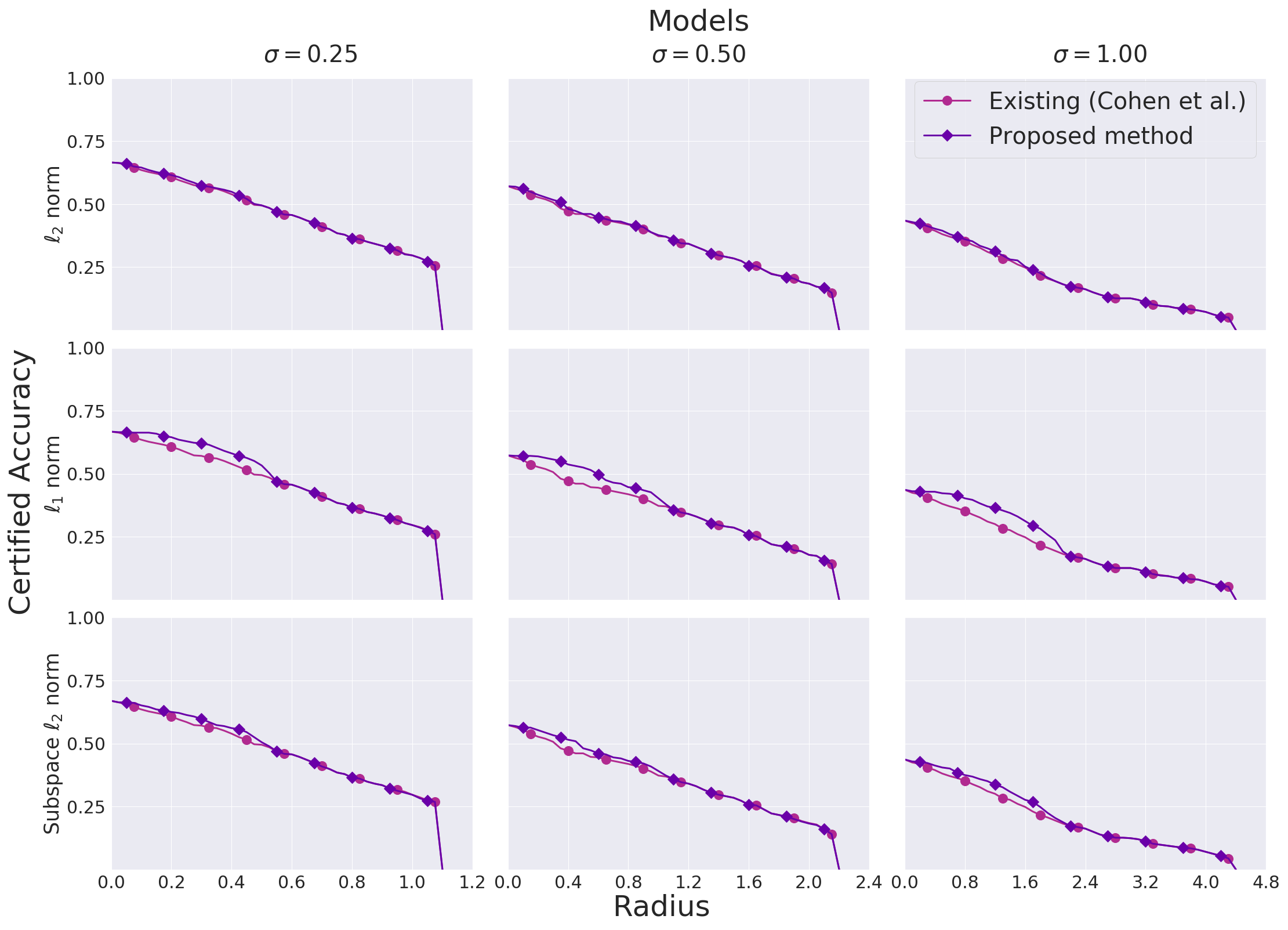}
    \caption{Certified Accuracy for Imagenet seen under various threat models and $\sigma$ values. The scale of $x$-axis is different for the 3 different models (denoted by training noise variance) as the certified radii we get for these three models have different ranges.}
    \label{fig: ImagenetCertAcc}
\end{figure}
The findings here are similar to the ones reported for CIFAR. As expected we see from Figure \ref{fig: ImagenetCertAcc} that the smallest improvements are for $\ell_2$ norm threat model where the new framework gives only marginal improvement over the $\ell_2$ radius certified by existing methods. This follows from the fact that the existing methods already produce near-optimal certified $\ell_2$ radii. However, certifying a significantly bigger certified safety region allows us to give significant improvements over the certified $\ell_1$ radius and the subspace $\ell_2$ radii (the subspace considered here is the red channel of the image, i.e., we only allow perturbations over red component of the RGB pixels of the image). 

From these figures we are also able to see that, for any given model, most of the improvement in certified accuracy occurs at smaller values of radius $R$. We think one of the causes for this is the interval size of our estimates of $y^{(0)}, y^{(1)}$. So, we give the following results using larger number of samples to estimate both $y^{(0)}, y^{(1)}$. 
\begin{figure}[ht]
    \centering
    \includegraphics[width=\textwidth]{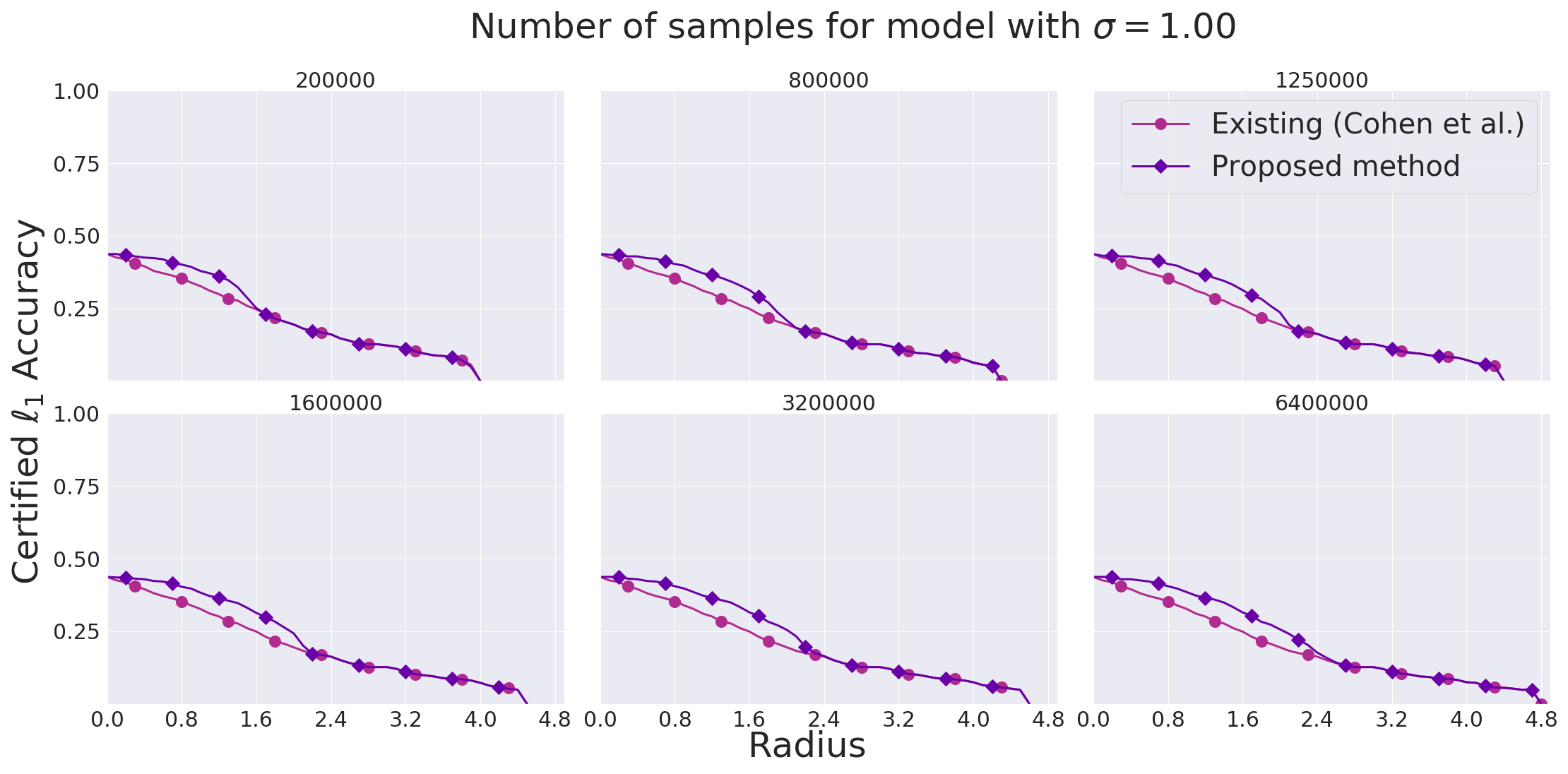}
    \caption{Effect of number of samples used on certified accuracy.}
    \label{fig: SamplesCertAcc}
\end{figure}

In Figure \ref{fig: SamplesCertAcc} we show the results by using the observed values from experiments using $N_{obs} = 1250000$ and then constructing the estimates assuming we used $N = 200000, 800000, 1250000, 1600000, 3200000, 6400000$ samples respectively. Using these estimates we get a certified $\ell_1$ radius for both the existing method and for the proposed method. We see that using larger number of samples allows us to get improvements at even larger values of $R$. However, we note that it is still not possible to get improvements at very high values of $R$. We think this would require very precise bounds for $y^{(0)}, y^{(1)}$ and thus a very high number of samples.
\end{document}